\pdfoutput=1
\documentclass[10pt,twocolumn,letterpaper]{article}

\usepackage{cvpr}
\usepackage{times}
\usepackage{epsfig}
\usepackage{graphicx}
\usepackage{amsmath}
\usepackage{amssymb}

\usepackage{booktabs}
\usepackage{placeins}
\usepackage{latexsym} 
\usepackage{subfigure}
\usepackage{url}
\usepackage{color} 
\usepackage{overpic}
\usepackage[table]{xcolor}
\usepackage{rotating}

\usepackage{multirow}
\usepackage[ruled,vlined]{algorithm2e}
\SetCommentSty{mycommfont}

\usepackage[pagebackref=true,breaklinks=true,letterpaper=true,colorlinks,bookmarks=false]{hyperref}

\cvprfinalcopy 

\def\cvprPaperID{16} 

\ifcvprfinal\fi

\newcommand{\vrho}{\ensuremath{\boldsymbol{\rho}}}

\begin{document}

\title{Color Constancy Using CNNs}

\author{Simone Bianco$^1$ \qquad Claudio Cusano$^2$ \qquad Raimondo Schettini$^1$\\
$^1$University of Milan-Bicocca, Italy\\
$^2$University of Pavia, Italy\\
{\tt\small \{simone.bianco,raimondo.schettini\}@disco.unimib.it, claudio.cusano@unipv.it }
}

\maketitle

\begin{abstract}
In this work we describe a Convolutional Neural Network
(CNN) to accurately predict the scene illumination. Taking image patches as input, the CNN
works in the spatial domain without using hand-crafted features
that are employed by most previous methods. The network
consists of one convolutional layer with max 
pooling, one fully connected layer and three output nodes.
Within the network structure, feature learning and regression
are integrated into one optimization process, which
leads to a more effective model for estimating scene illumination.
This approach achieves state-of-the-art performance on a standard dataset of RAW images. 
Preliminary experiments on images with spatially varying illumination demonstrate the stability of the local illuminant estimation
ability of our CNN.
\end{abstract}

\section{Introduction}

Many computer vision problems in both still images and videos can make use of color constancy processing as a pre-processing step to make sure that the recorded color of the objects in the scene does not change under different illumination conditions.
The observed color of the objects in the scene depends on the intrinsic color of the object (i.e. the surface spectral reflectance), on the illumination, and on their relative positions.

In general there are two methodologies to obtain reliable color description
from image data: computational color constancy and color invariance \cite{lee2014taxonomy}.
Computational color constancy is a two-stage operation: the former is specialized on estimating the color of the scene illuminant from the image data, the latter corrects the image on the basis of this estimate to generate a new image of the scene as if it was taken under a reference light source. 
Color invariance methods instead represent images by features which remain unchanged with respect to specific imaging condition.


In this work we focus on computation color constancy, using a CNN to learn discriminant features
for the illuminant estimation task. Recently, deep neural networks
have gained the attention of numerous researchers outperforming state-of-the-art approaches on various computer vision tasks \cite{kavukcuoglu2010learning,krizhevsky2012imagenet}. 
One of CNN's advantages is that it can take raw images as input and incorporate feature learning into the training process. With a deep structure,
CNN can learn complicated mappings while requiring minimal domain knowledge.

To the best of our knowledge, this is the first work that investigates the use of CNNs for illuminant estimation. 
The main contribution of our paper is that we propose a
novel method that allows learning and prediction of scene
illuminant on local regions. Previous approaches typically
accumulate features over the entire image to obtain statistics
for estimating the overall illuminant, and only a few approaches have shown
the ability to estimate spatially varying illuminant. By contrast, our method can estimate the illuminant 
on small patches (such as $32 \times 32$). 
We show experimentally that the proposed method advances
the state-of-the-art on a standard dataset of RAW images. 
In addition to the superior overall performance,
we also show quantitative and qualitative results that demonstrate the quality of the local
illuminant estimation of our method.

%
%
\section{{Problem formulation and related works}}
\label{sec:approach}
The image values for a Lambertian surface located at the pixel with coordinates $(x,y)$ can be seen as a function $\boldsymbol{\rho}(x,y)$, mainly dependent on three physical factors: the illuminant spectral power distribution $I(x,y,\lambda)$, the surface spectral reflectance $S(x,y,\lambda)$ and the sensor spectral sensitivities $\mathbf{C}(\lambda)$. Using this notation $\boldsymbol{\rho}(x,y)$ can be expressed as
\begin{equation}
  \boldsymbol{\rho}(x,y) = \int  I(x,y,\lambda) S(x,y,\lambda) \mathbf{C}(\lambda) \mathrm{d}\lambda,
\label{eq:colorformationequation}
\end{equation}
{where $\lambda$ is the wavelength}, $\boldsymbol{\rho}$ and $\mathbf{C}(\lambda)$ are three-component vectors and the integration is performed over the visible spectrum. 
The goal of color constancy is to estimate the color $\mathbf{I}(x,y)$ of the scene illuminant, i.e. the projection of $I(x,y,\lambda)$ on the sensor spectral sensitivities $\mathbf{C}(\lambda)$:
\begin{equation}
  \mathbf{I}(x,y) = \int  I(x,y,\lambda) \mathbf{C}(\lambda) \mathrm{d}\lambda.
\end{equation}
{
{Usually the illuminant color is estimated up to a scale factor as it is more important to estimate the chromaticity of the scene illuminant than its overall intensity \cite{HorFin04}}. 
}
Since the only information available are the sensor responses $\boldsymbol{\rho}$ across the image, color constancy is an under-determined problem \cite{FunBarMar98} and thus further assumptions and/or knowledge are needed to solve it. 

{
Several computational color constancy algorithms have been proposed, each based on different assumptions. The most common assumption made a uniform  light source color across the scene, i.e. $\mathbf{I}(x,y)=\mathbf{I}$.} 

State-of-the-art solutions can be divided into two main classes: statistic approaches, and learning-based
approaches. Statistic approaches estimate the scene illumination only on the base of the content in a single
image making assumptions about the nature of color images exploiting statistical or physical
properties; learning-based approaches require training
data in order to build a statistical image model, prior to estimation of illumination. 

\subsection{Statistic-based algorithms}
Van~de~Weijer et al. \cite{vandeWGev07} have unified a variety of algorithms. These algorithms estimate the illuminant color $\mathbf{I}$ by implementing instantiations of the following equation:
\begin{equation}
  \label{eq:general-method}
  \mathbf{I}(n, p, \sigma) = \frac{1}{k} \left( \iint \left| \nabla^n
  \boldsymbol{\rho}_\sigma(x,y) \right|^p \mathrm{d}x \ \mathrm{d}y
  \right)^{\frac{1}{p}},
\end{equation}
where $n$ is the order of the derivative, $p$ is the Minkowski norm, $\boldsymbol{\rho}_\sigma(x,y) = \boldsymbol{\rho}(x,y) \otimes G_\sigma(x,y)$ is the convolution of the image with a Gaussian filter $G_\sigma(x,y)$ with scale parameter $\sigma$, and $k$ is a constant to be chosen such that the illuminant color $\mathbf{I}$ has unit length ({using the $2-$norm}). The integration is performed over all pixel coordinates. Different $(n,p,\sigma)$ combinations correspond to different illuminant estimation algorithms, each based on a different assumption. For example, the Gray World algorithm \cite{GW} -- generated setting $(n,p,\sigma)=(0,1,0)$ -- is based on the assumption that the average color in the image is gray and that the illuminant color can be estimated as the shift from gray of the averages in the image color channels; the White Point algorithm \cite{WP} -- generated setting $(n,p,\sigma)=(0,\infty,0)$ -- is based on the assumption that there is always a white patch in the scene and that the maximum values in each color channel are caused by the reflection of the illuminant on the white patch, and they can be thus used as the illuminant estimation; the Gray Edge algorithm \cite{vandeWGev07} -- generated setting for example $(n,p,\sigma)=(1,0,0)$ -- is based on the assumption that the average color of the edges is gray and that the illuminant color can be estimated as the shift from gray of the averages of the edges in the image color channels. \\
The Gamut Mapping assumes that for a given illuminant, one observes only a limited gamut of colors \cite{gamutmapF}. It has a preliminary phase in which a canonical illuminant is chosen and the canonical gamut is computed observing as many surfaces under the canonical illuminant as possible. Given an input image with an unknown illuminant, its gamut is computed and the illuminant is estimated as the mapping that can be applied to the gamut of the input image, resulting in a gamut that lies completely within the canonical gamut and produces the most colorful scene. If the spectral sensitivity functions of the camera are known, the Color by Correlation approach could be also used \cite{CbC}. \\

\subsection{Learning-based algorithms}
The learning-based color constancy algorithms, that estimate the scene illuminant using a model that is learned on training data, can be subdivided into two main subcategories: probabilistic methods and fusion/selection based methods. 
Bayesian approaches \cite{cambridge} model the variability of reflectance and of illuminant as random variables, and then estimate illuminant from the posterior distribution conditioned on image intensity data. \\
Given a set computational color constancy algorithms, in \cite{BiaCioCusSch08} an image classifier is trained to classify the images as indoor and outdoor, and different experimental frameworks are proposed to exploit this information in order to select the best performing algorithm on each class.
In \cite{BiancoPR} it has been shown how intrinsic, low level properties of the images can be used to drive the selection of the best algorithm (or the best combination of algorithms) for a given image. The algorithm selection and combination is made by a decision forest composed of several trees on the basis of the values of a set of heterogeneous features. \\
In \cite{NIS2011} the Weibull parametrization has been used to train a maximum likelihood classifier based on mixture of Gaussians to select the best performing color constancy method for a certain image. \\
In \cite{chzcc2011} a statistical model for the spatial distribution of colors in white balanced images is developed, and then used to infer illumination parameters as those being most likely under their model.
High level visual information has been used to select the best illuminant out of a set of possible illuminants \cite{hilevelinfo}. This is achieved by restating the problem in terms of semantic interpretability of the image.
Several color constancy methods are applied to generate a set of illuminant hypotheses. For each illuminant hypothesis, they correct the image, evaluate the likelihood of the semantic content of the corrected image, and select the most likely illuminant color. 
In \cite{bianco2012color,bianco2014adaptive} the use of automatically detected objects having intrinsic color is investigated. In particular, it is investigated how illuminant estimation can be performed exploiting the color statistics extracted from the faces automatically detected in the image. 
When no faces are detected in the image, any other algorithm in the state-of-the-art can be used.  
In \cite{joze2012exemplar,joze2014exemplar} the surfaces in the image are exploited and the color constancy problem is addresses by unsupervised learning of an appropriate model for each training surface in training images. The model for each surface is defined using both texture features and color features.
In a test image the nearest neighbor model is found for each surface and its illumination is estimated  by comparing the statistics of pixels belonging to nearest neighbor surfaces  and the target surface. The final illumination estimation results from combining these estimated illuminants over surfaces to generate a unique estimate.

\section{{The proposed approach}}
\label{sec:method}
The proposed framework of using CNN for illuminant 
estimation is as follows. Given color image, we sample non-overlapping
patches from it and for each of them we perform a contrast normalization through histogram stretching. We use a CNN to estimate the
illuminant for each patch and combine the patch scores to obtain an illuminant estimation for the image.


\subsection{Network architecture}
\begin{figure*}
	\centering
		\includegraphics[width=1.45\columnwidth]{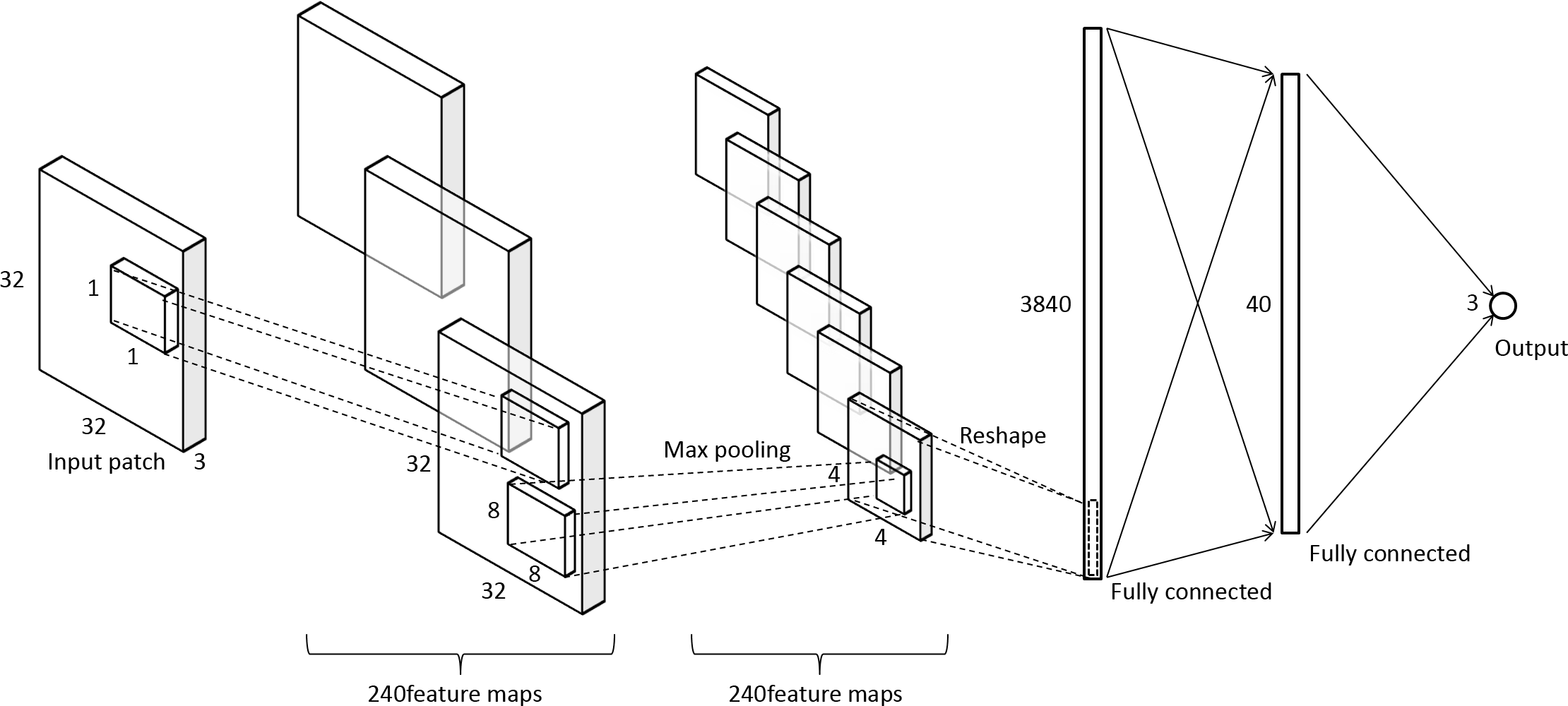}
	\caption{The architecture of our CNN.}
	\label{fig:architecture}
\end{figure*}

The proposed network consists of five layers. Figure \ref{fig:architecture}
shows the architecture of our network, which is a 32x32x3 - 32x32x240 - 4x4x240 - 40 - 3 structure. The
input is contrast normalized 32x32 image patches. The first
layer is a convolutional layer which filters the input with
240 kernels each of size 1x1x3 with a stride of 1 pixel. The
convolutional layer produces 240 feature maps each of size
32x32, followed by a max-pooling operation with 8x8 kernels and stride of 8 pixels that reduces each
feature map to a 4x4 feature map. These are reshaped into a 3840 (4x4x240) vector. One fully connected
layer of 40 nodes come after the reshaping. The last
layer is a simple linear regression with a three dimensional
output that gives the illuminant estimate.

\subsection{Contrast normalization}
In order to be robust across lighting condition and since in color constancy the illuminant has to be estimated up to a scale factor, all the extracted patches are contrast normalized. Among the different contrast enhancement techniques we have chosen the global histogram stretching as does not change the relative contributions of the three color channels. 

\subsection{Pooling}
In the convolution layer, the contrast normalized image
patches are convolved with 240 filters and each filter generates
a feature map. We then apply max-pooling on each feature
map to reduce the filter responses to a lower dimension.
In contrast with object recognition scenario, where pooling tends to be performed on small neighborhoods, we observe that even in case of spatially varying illuminants, these are 
locally homogeneous, i.e. the same illuminant tends to be present on all the the locations of a 32x32 patch. This permits the use of larger pooling kernels.

\subsection{ReLU nonlinearity}
Instead of traditional sigmoid or tanh neurons, we use
Rectified Linear Units (ReLUs) \cite{nair2010rectified} in the fully connected
layer. Krizhevsky et al. demonstrated that  that ReLUs
enable the network to train several times faster compared
to using tanh units while achieving almost identical performance \cite{krizhevsky2012imagenet}.

\subsection{CNN features}
Together with learning an ad-hoc CNN for the color constancy problem, we also investigate how a pre-trained one works on this problem. To this end, we extract a 4096-dimensional feature vector from each image using the Caffe \cite{jia2014caffe} implementation of the deep CNN described by Krizhevsky et al. \cite{krizhevsky2012imagenet}.  
Features are computed by forward propagation of a mean-subtracted $227 \times 227$ RGB RAW image through five convolutional layers and two fully connected layers. More details about the network architecture can be found in \cite{krizhevsky2012imagenet,jia2014caffe}.
The CNN was discriminatively trained on a large dataset (ILSVRC 2012) with image-level annotations to classify images into 1000 different classes. Features are obtained by extracting activation values of the last hidden layer. The extracted features are then used as input to a linear Support Vector Regressor (SVR) \cite{SVR} to estimate the illuminant color for each image. We refer to this method as AlexNet+SVR in the experimental results.

\section{Experimental Setup}
\label{sec:setup}
The aim of this section is to investigate if the proposed algorithm can outperform state-of-the-art algorithms in the illuminant estimation on a standard dataset of RAW images. 

\subsection{Image Datasets and Evaluation Procedure}
\begin{figure}
	\centering
		\includegraphics[width=\columnwidth]{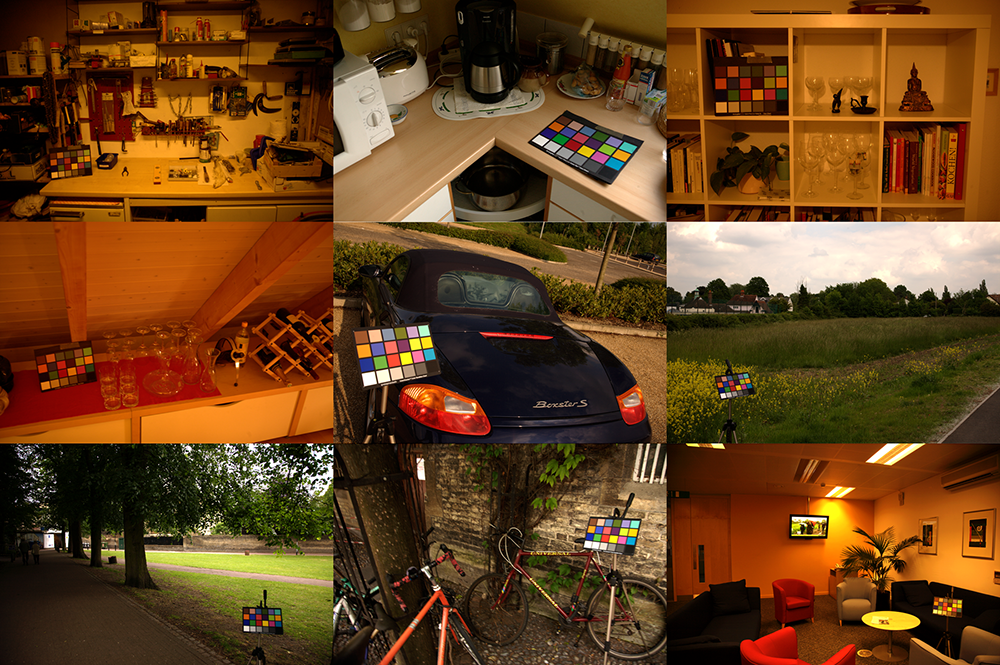}
	\caption{Example of images within the RAW dataset.}
	\label{fig:dataset}
\end{figure}

To test the performance of the proposed algorithm, a standard dataset of RAW camera images having a known color target are used. It is captured using high-quality digital SLR cameras in RAW format, and is therefore free of any color correction. 
The dataset \cite{cambridge} was originally available in sRGB-format, but Shi and Funt \cite{Shi} reprocessed the raw data to obtain linear images with a higher dynamic range (14 bits as opposed to standard 8 bits).
The dataset has been acquired using a Canon 5D and a Canon 1D DSLR cameras and consists of a total of 568 images. 
The Macbeth ColorChecker (MCC) chart is included in every scene acquired, and this allows to accurately estimate the actual illuminant of each acquired image. 
Examples of images within the RAW  
dataset are reported in Figure \ref{fig:dataset}.

\subsection{Error metric}
The error metric considered, as suggested by Hordley and Finlayson \cite{HorFin04}, is the angle between the RGB triplet of estimated illuminant ($\vrho_w$) and the RGB triplet of the measured ground truth illuminant ($\hat{\vrho_w}$):
\begin{equation}
  e_{\text{ANG}} = \arccos \left( \frac{\vrho^T_w \hat{\vrho_w}}{\| \vrho_w \|  \| \hat{\vrho_w}\| } \right) .
\end{equation}

\subsection{Benchmark algorithms}
\label{sec:bench}

Different benchmarking algorithms for color constancy are considered. Since each image of the dataset contains only one MCC, only global color constancy algorithms based on the assumption of uniform illumination can be compared. Six of them are generated varying the three variables $(n,p,\sigma)$ in Equation~\ref{eq:general-method}, and correspond to well known and widely used color constancy algorithms. 
The values chosen for $(n,p,\sigma)$ are reported in Table \ref{tab:parametri} and set as in \cite{ginseng11}. 
The algorithms are used in the original authors' implementation which is freely available online (\url{http://lear.inrialpes.fr/people/vandeweijer/code/ColorConstancy.zip}). The seventh algorithm is the pixel-based Gamut Mapping \cite{gamutmappingGevers}. The value chosen for $\sigma$ is also reported in Table \ref{tab:parametri}.
{
The other algorithms considered are illumination chromaticity estimation via Support Vector Regression (SVR \cite{funt2004estimating}); the Bayesian (BAY \cite{cambridge}); the Natural Image Statistics (NIS \cite{NIS2011}); the High Level Visual Information \cite{hilevelinfo}: bottom-up (HLVI BU), top-down (HLVI TD), and their combination (HLVI BU\&TD); the Spatio-Spectral statistics \cite{chzcc2011}: with Maximum Likelihood estimation (SS ML), and with General Priors (SS GP); the Automatic color constancy Algorithm Selection (AAS) \cite{BiancoPR} and the Automatic Algorithm Combination (AAC) \cite{BiancoPR}; the Exemplar-Based color constancy (EB) \cite{joze2012exemplar}; the Face-Based (FB) color constancy algorithm \cite{bianco2012color} using GM or SS ML when no faces are detected.} 

\begin{table}[!ht]
\caption{{Values chosen for $(n,p,\sigma)$ for the state-of-the-art algorithms which are instantiations of Eq.\ref{eq:general-method}.}}
\label{tab:parametri}
\centering
\begin{tabular}{lrrr}
  \toprule
  Algorithm & $n$ & $p$ &  $\sigma$ \\
  \midrule
Gray World (GW)     							&   0 &   1          &  0 \\
White Point (WP)     							&   0 &   $\infty $  &  0 \\
Shades of Gray (SoG)    					&   0 &   4          &  0 \\
general Gray World (gGW)    		  &   0 &   9          &  9 \\
1st-order Gray Edge (GE1)    			&   1 &   1          &  6 \\
2nd-order Gray Edge (GE2)    			&   2 &   1          &  1  \\
Gamut Mapping (GM)								&   0 &   0          &  4  \\
	\bottomrule
\end{tabular}
\end{table}

The last algorithm considered {is} the Do Nothing (DN) algorithm which gives the same estimation for the color of the illuminant ($\mathbf{I}=[1 \ 1 \ 1]$) for every image, i.e. it assumes that the image is already correctly balanced.

\subsection{CNN learning}
We train our CNN on 32x32 random patches taken from images in RAW format. Images have been resized to $\max (w,h)=1200$. The net is learned using a thee-fold cross validation on the folds provided with the dataset: for each run one is used for training, one for validation and the remaining one for test. For training, we assign each patch with the illuminant groundtruth associated to the image to which it belongs. At testing time, we generate a single illuminant estimation per image by pooling the the predicted patch illuminants. 
By taking image patches as input, we have a much larger number of training samples
compared to using the whole image on a given dataset, which particularly meets the needs of CNNs.
Net parameters have been learned using Caffe \cite{jia2014caffe} with euclidean loss. 
{The learned net is then fine-tuned by using as loss the angular error and adding knowledge about the way local estimates are pooled to generate a single global estimate for each image.}

\section{Results and Discussion}
\label{sec:results}
In Table \ref{tab:errori} the minimum, the 10$^{th}$-percentile, the median, the average, the 90$^{th}$-percentile, and the maximum of the angular errors obtained by the considered state-of-the-art algorithms and the proposed approach on the {RAW dataset} are reported. 
The table is divided into three blocks and for each of them the best result for each statistic is reported in bold. The first block includes statistic-based algorithms, the second one learning-based algorithms, and the third one the different variants of the proposed approach.  

From the results it is possible to see that the deep CNN pre-trained on ILSVRC 2012 \cite{krizhevsky2012imagenet} coupled with SVR (i.e. AlexNet+SVR) is already able to outperform most statistic-based algorithms and some learning-based ones. 
The next entry is the angular error made by our CNN on the patches. It is possible to see that, with respect to the median error, the proposed approach is able to outperform half of the learning-based algorithms considered. The next two entries are the results obtained by our approach by pooling patch-based illuminant estimations over the whole image. Two very simple pooling strategies have been considered, i.e. average and median pooling. It is possible to see that both outperform most of the state-of-the-art algorithms in terms of median error, with the average-pooling having a maximum error just 0.3\% worse than the best algorithm in the state-of-the-art.
The last entry reported is the fine-tuned CNN with median-pooling and angular error loss. We can notice that it reaches both a median, an average and a maximum angular error better than all the state-of-the-art algorithms considered. The improvement is of 1.5\%, 5.1\% and 0.2\% respectively but it is remarkable that they have been achieved by the same algorithm, while the best values in the state-of-the-art for the same statistics were obtained by three different learning-based algorithms. i.e. FB+GM, EB and SS GP respectively.

\begin{figure*}
	\centering
	\begin{tabular}{cccc}
		\includegraphics[width=0.40\columnwidth]{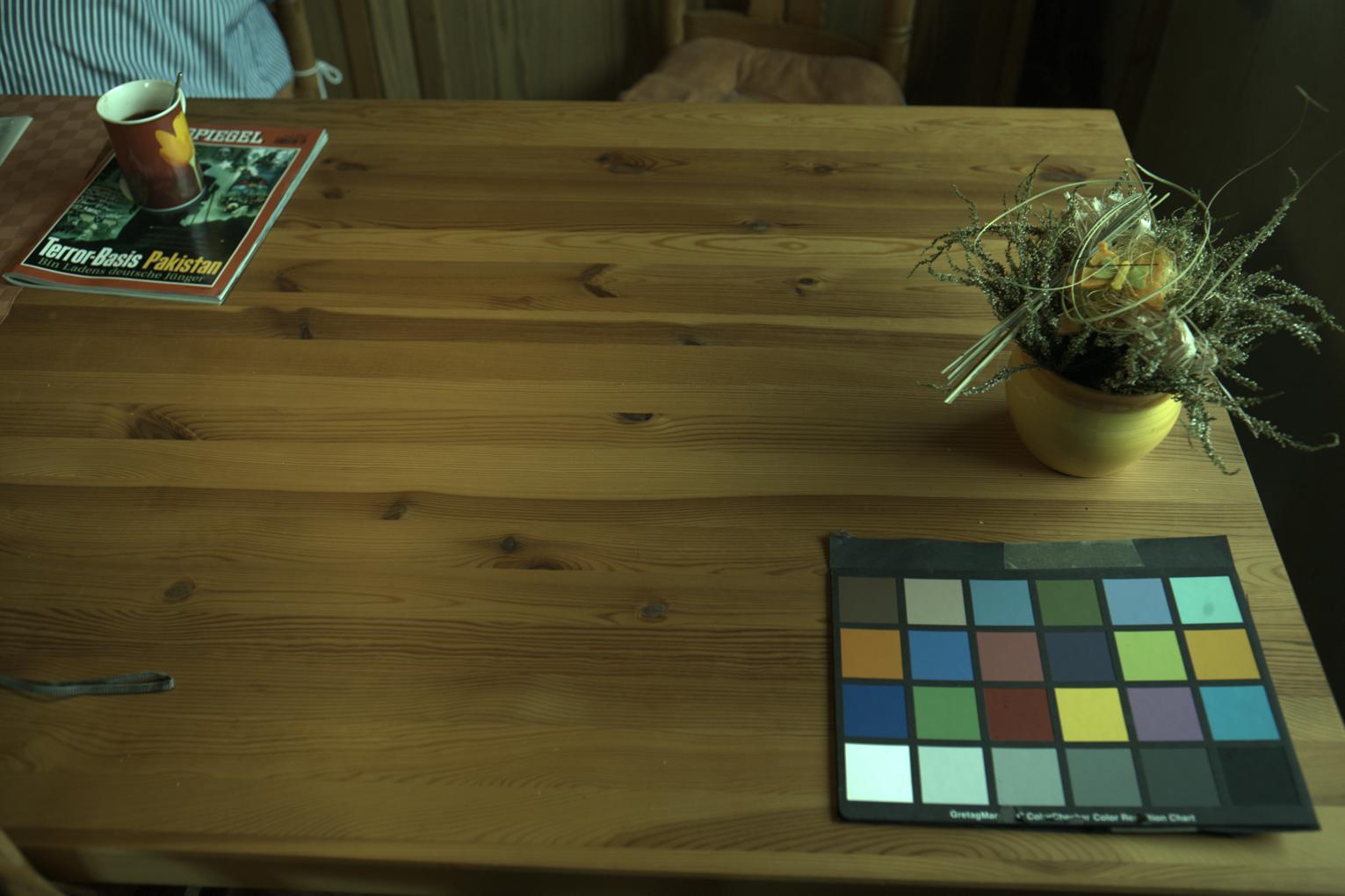} & \includegraphics[width=0.40\columnwidth]{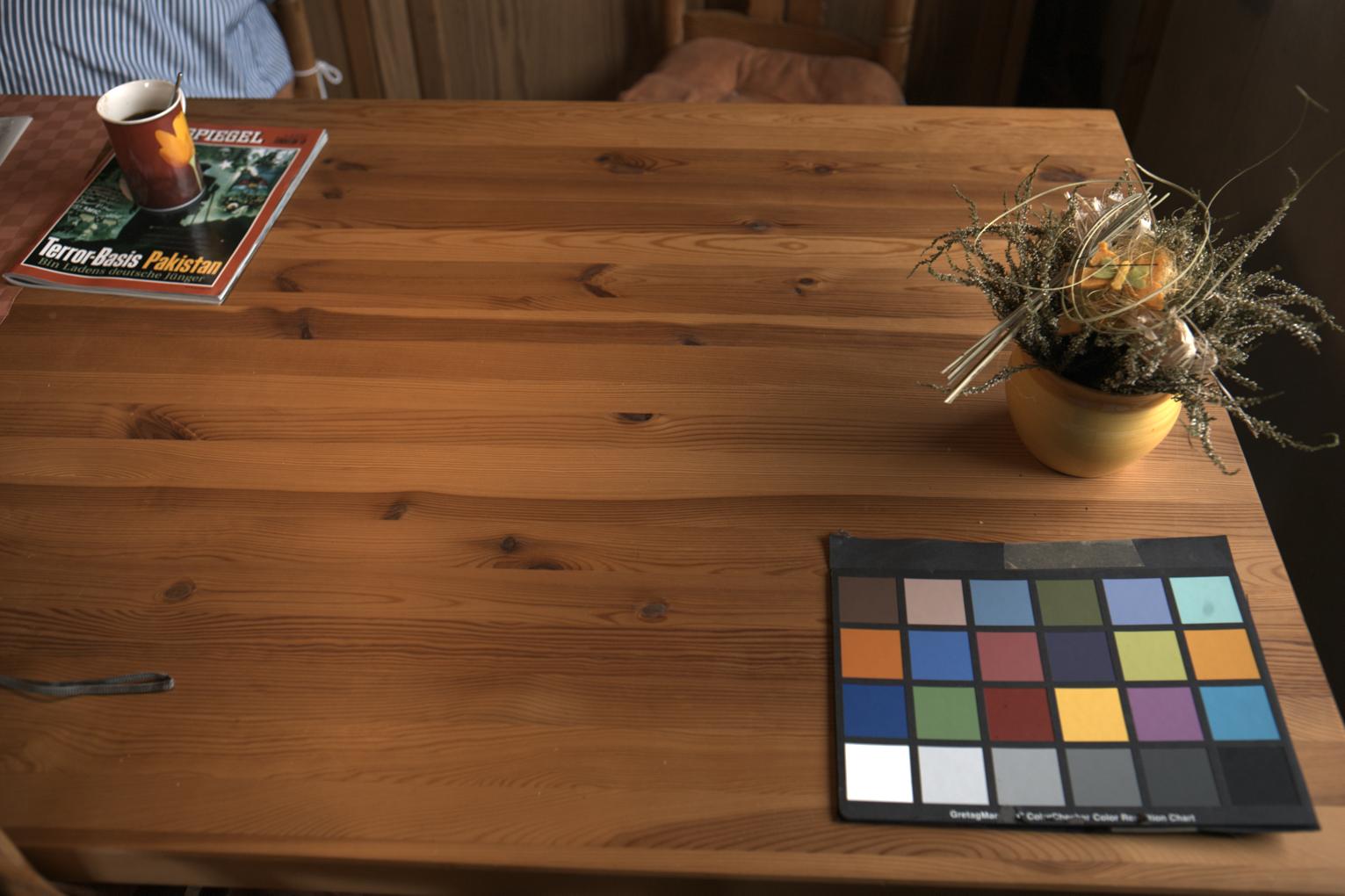} & \includegraphics[width=0.40\columnwidth]{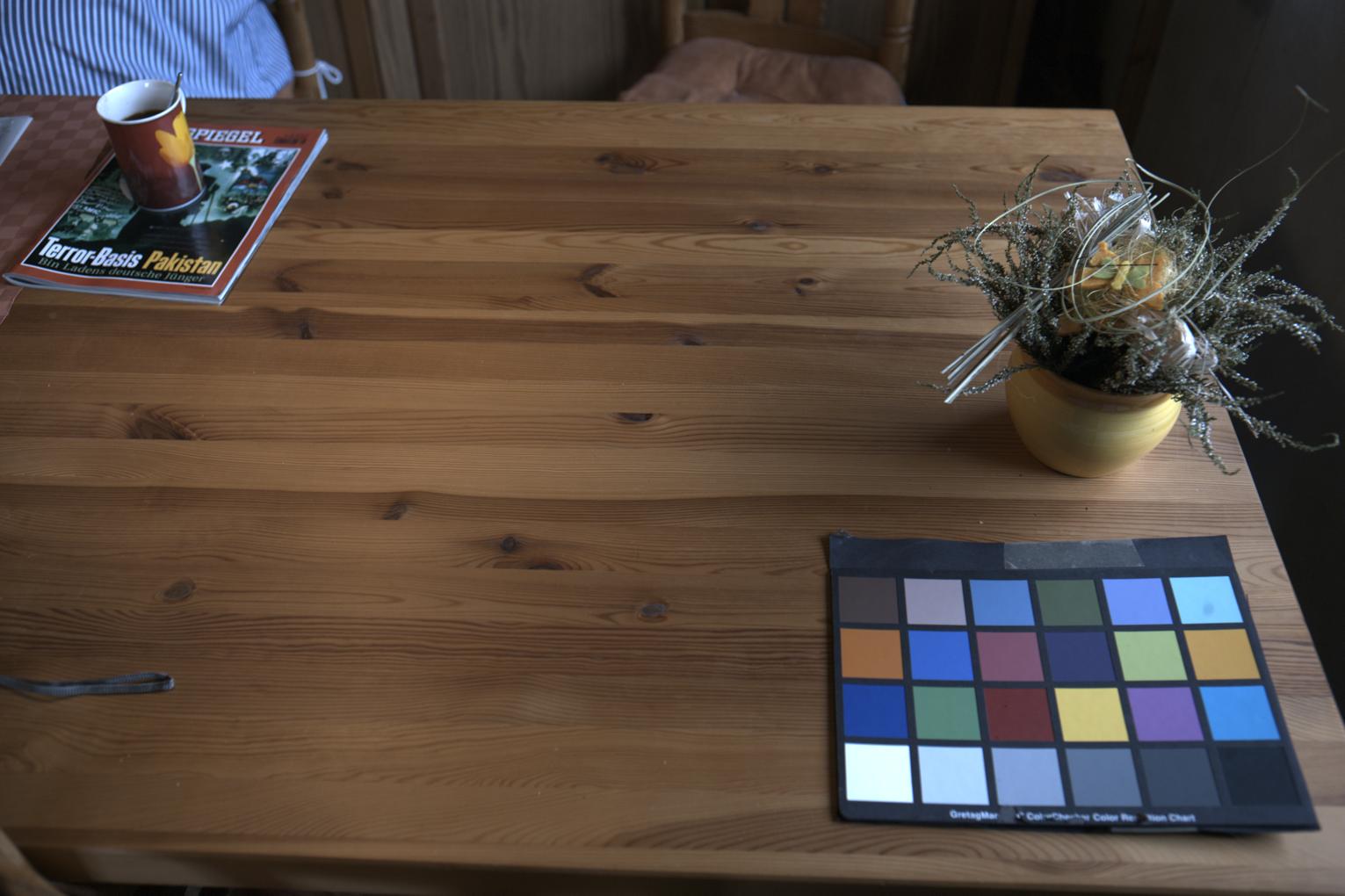} & \includegraphics[width=0.40\columnwidth]{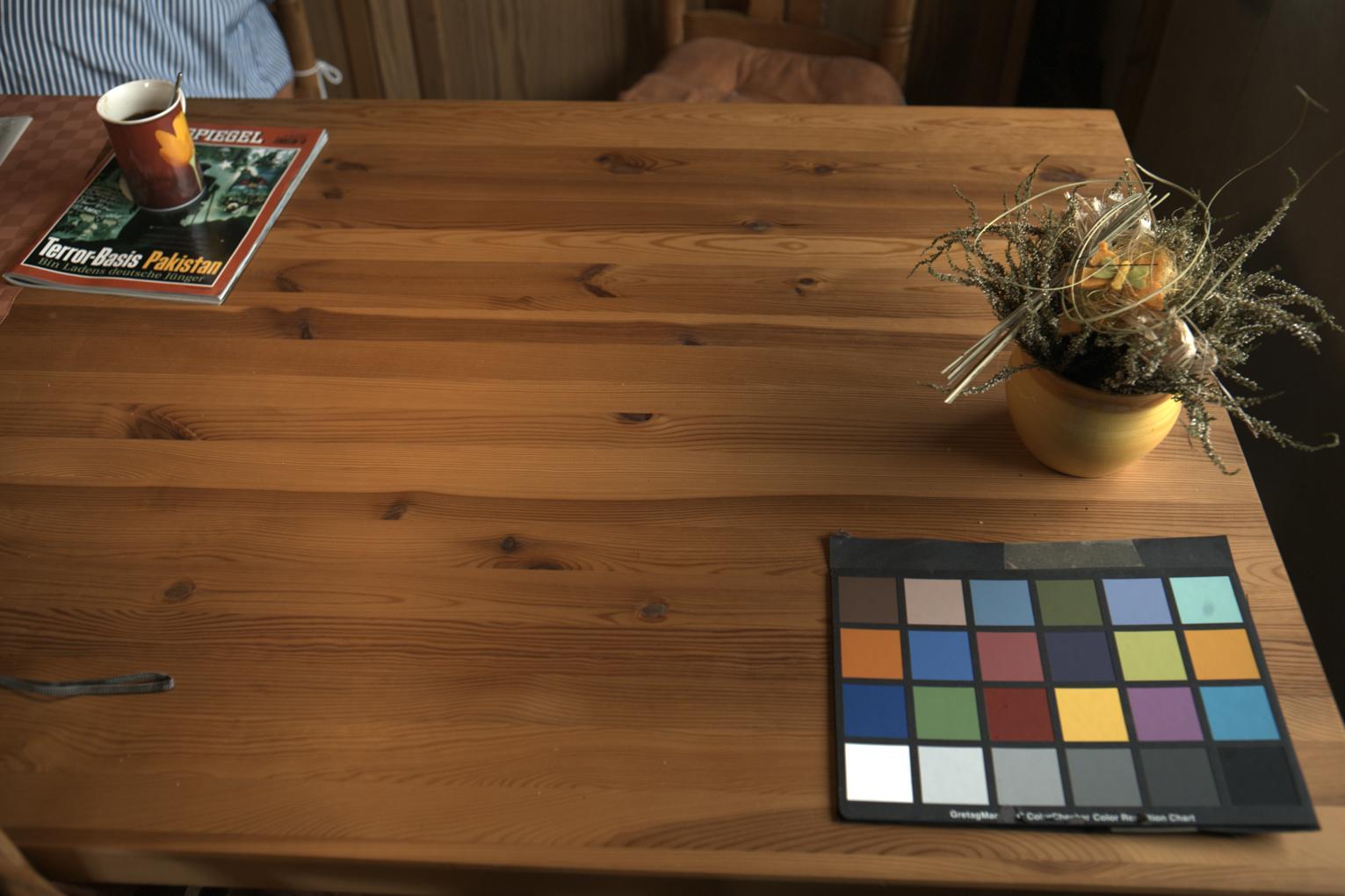} \\
		Original & groundtruth (0$^\circ$) & CNN fine-tuned (14.77$^\circ$) & AAS (1.48$^\circ$) \\
		

		\includegraphics[width=0.40\columnwidth]{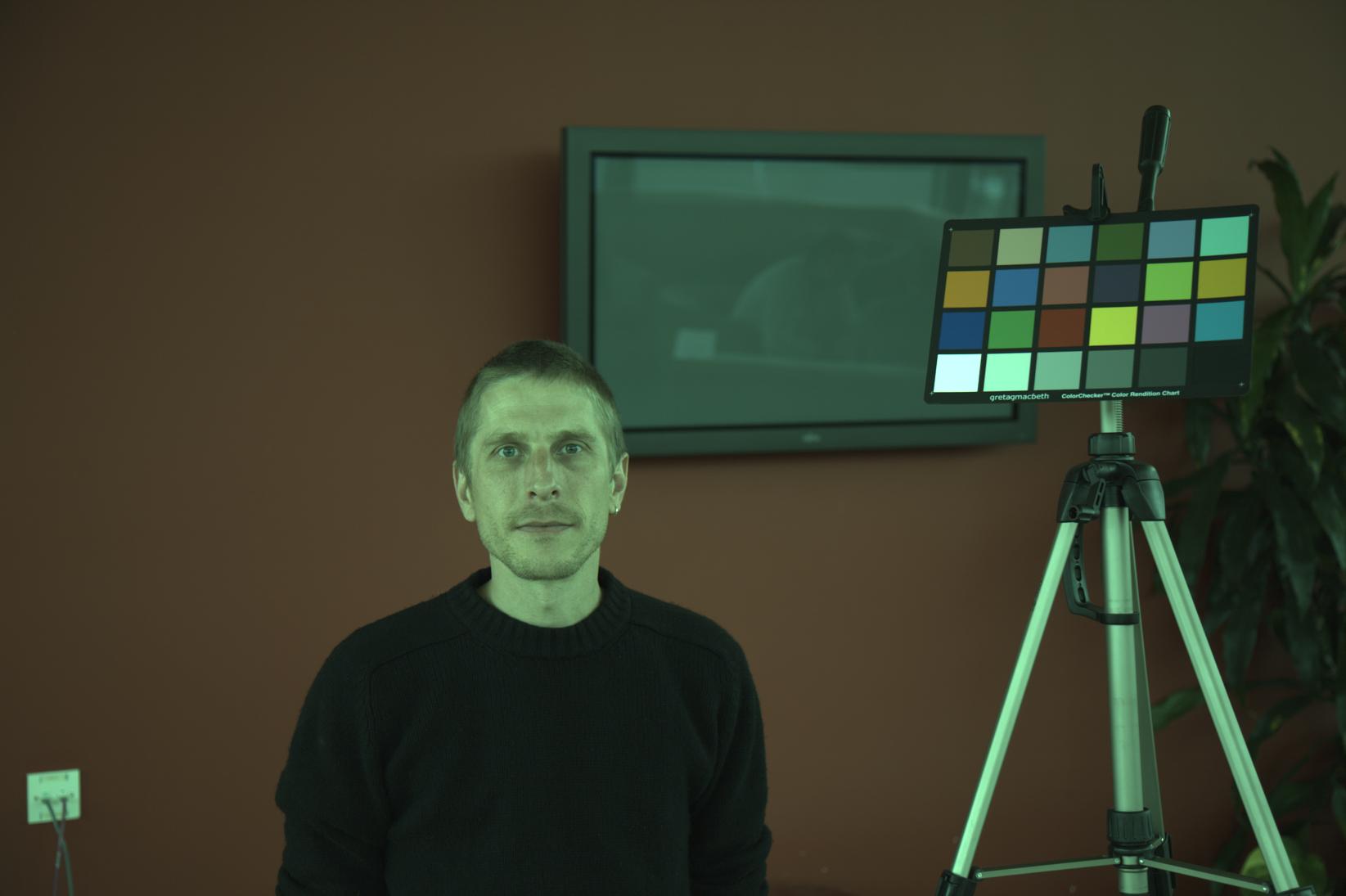} & \includegraphics[width=0.40\columnwidth]{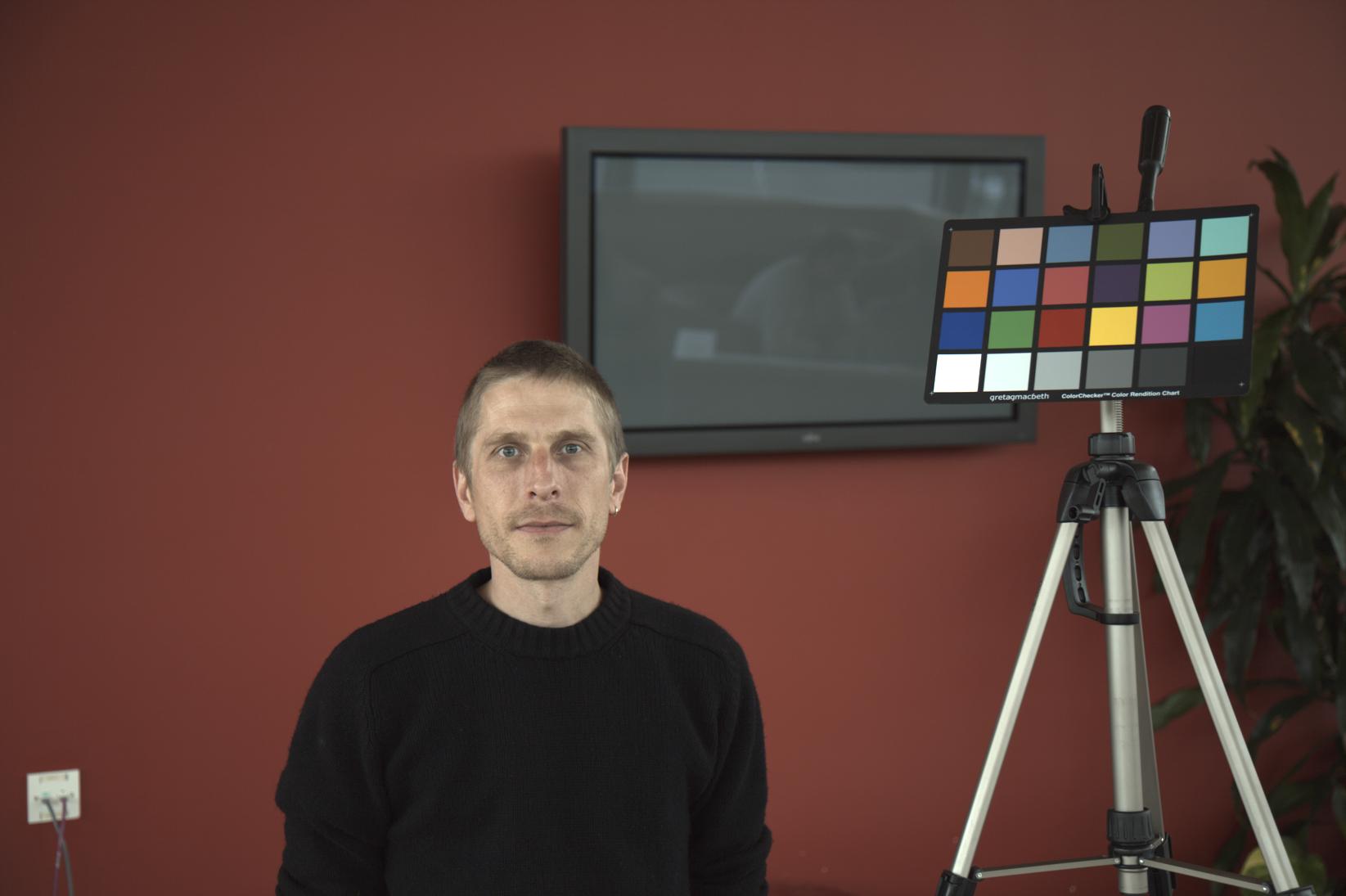} & \includegraphics[width=0.40\columnwidth]{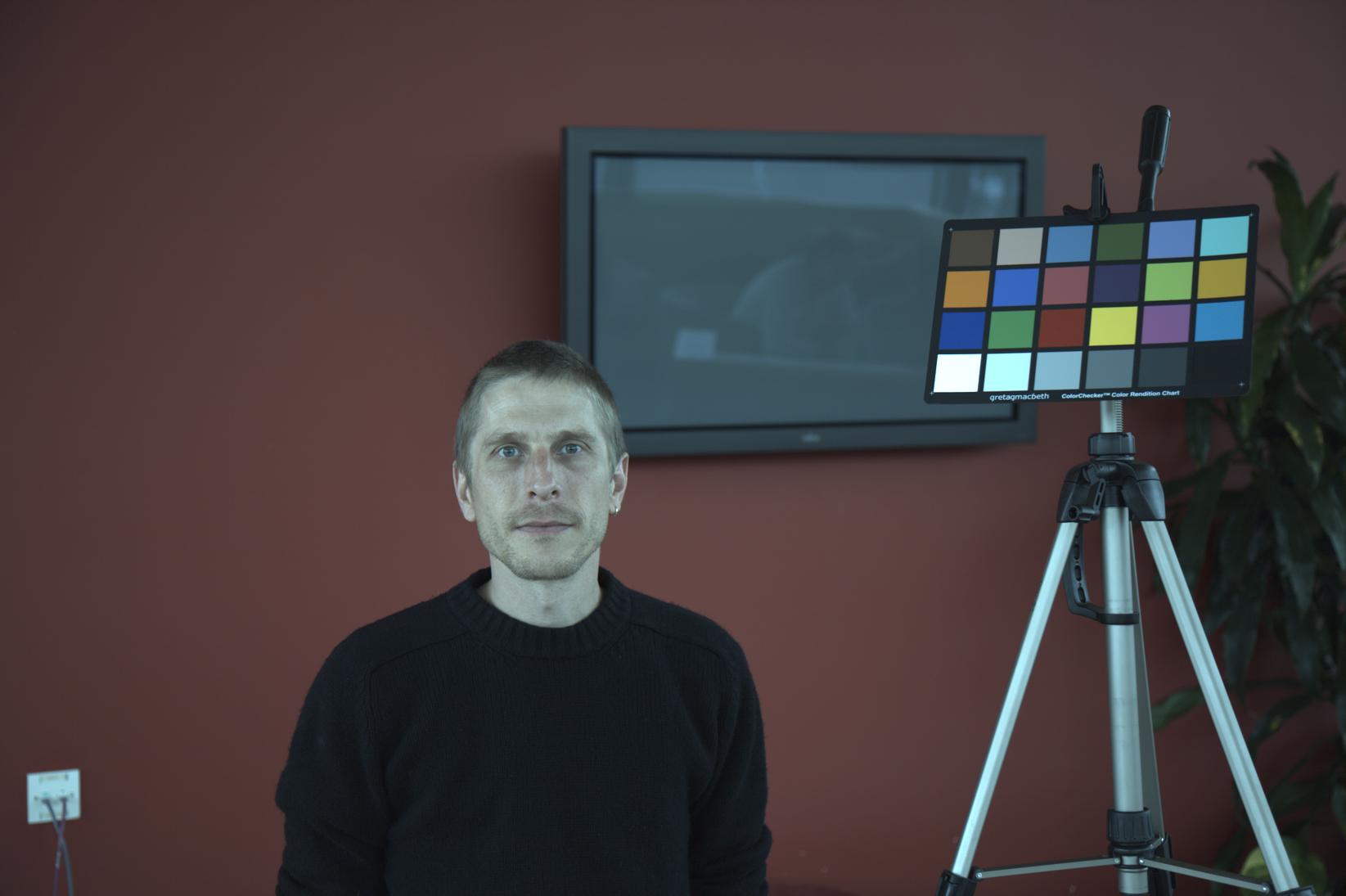} & \includegraphics[width=0.40\columnwidth]{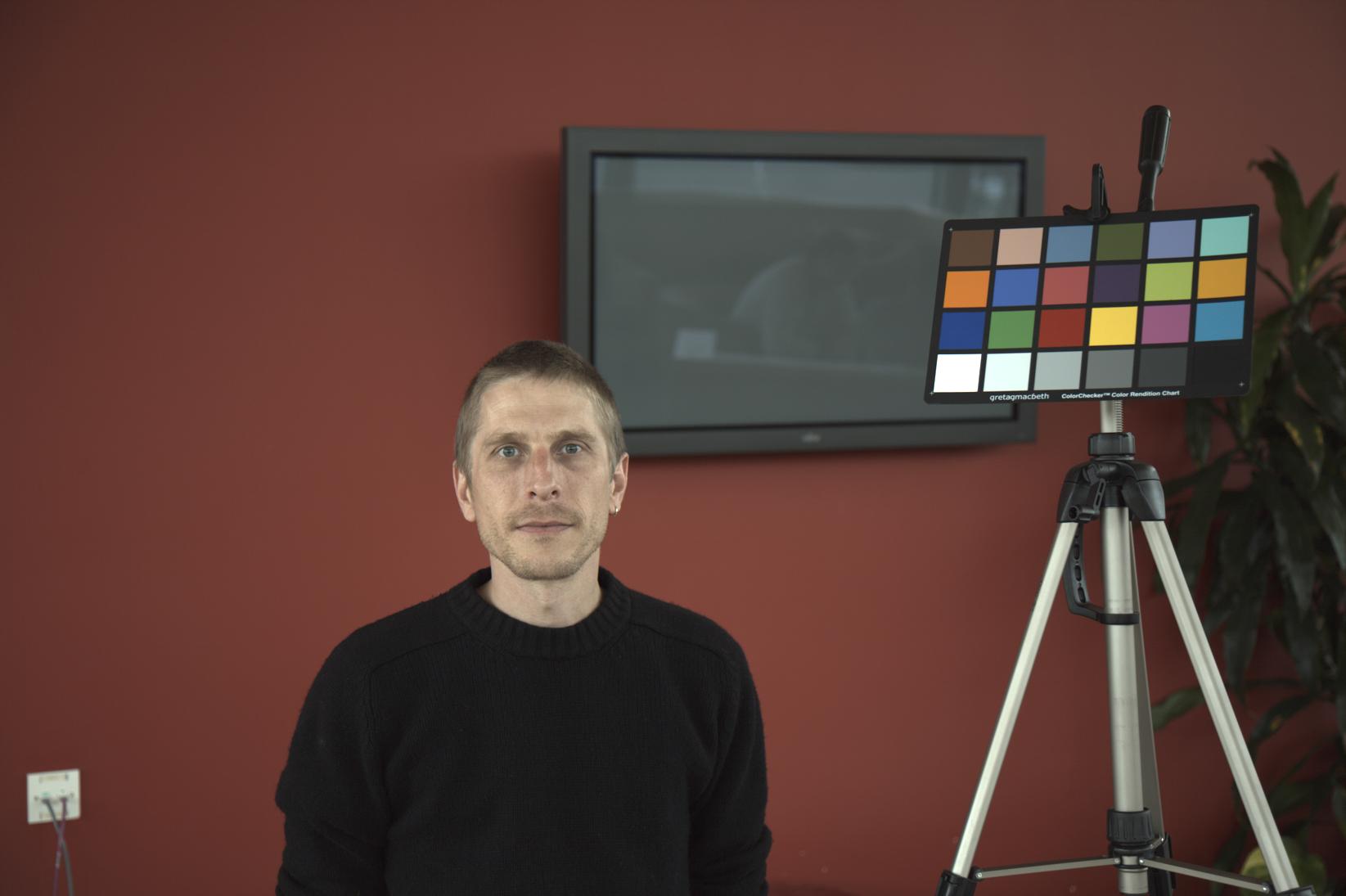}\\
		Original & groundtruth (0$^\circ$) & CNN fine-tuned (11.95$^\circ$) & FB+GM (0.27$^\circ$)\\




		\includegraphics[width=0.40\columnwidth]{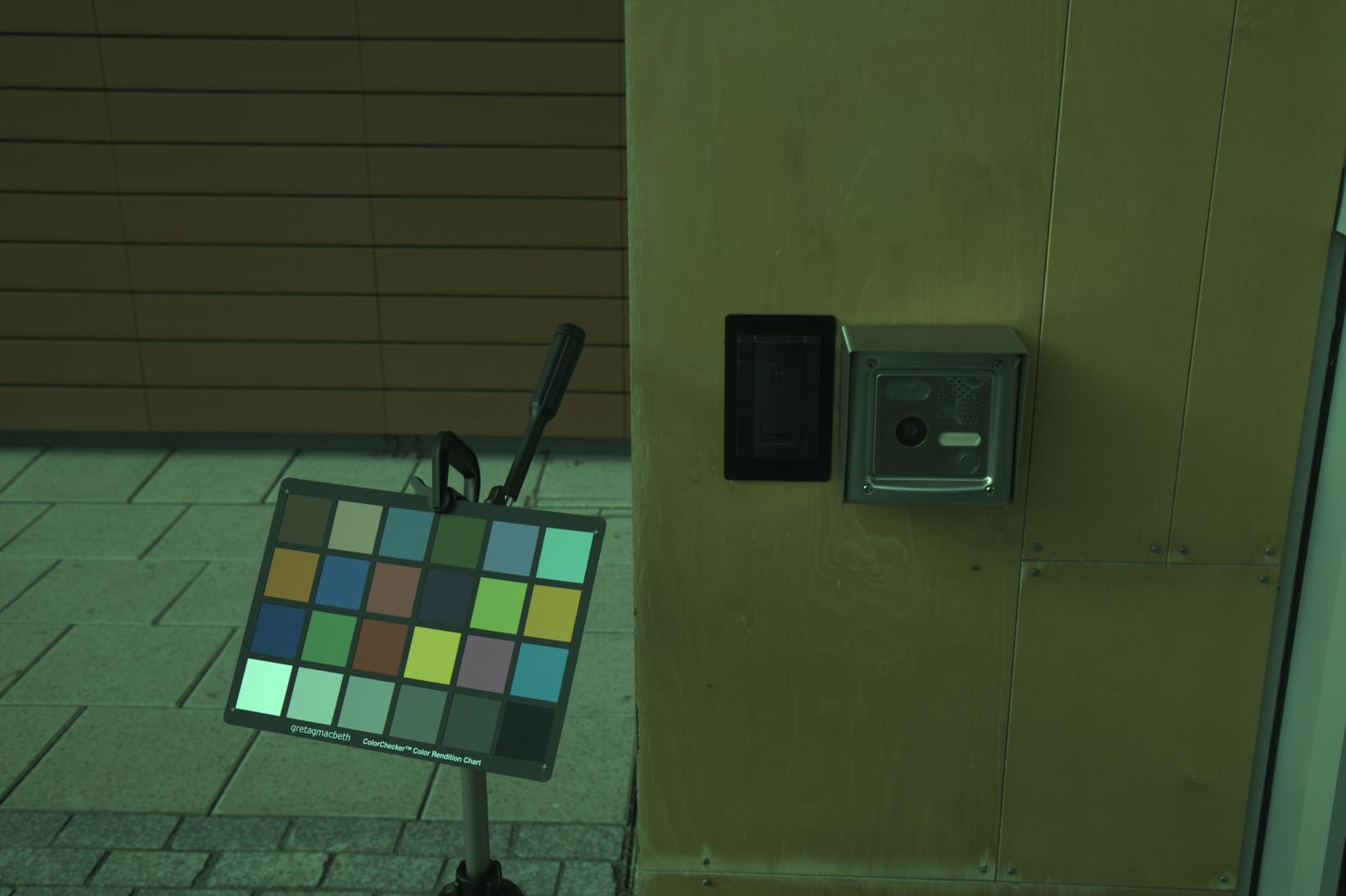} & \includegraphics[width=0.40\columnwidth]{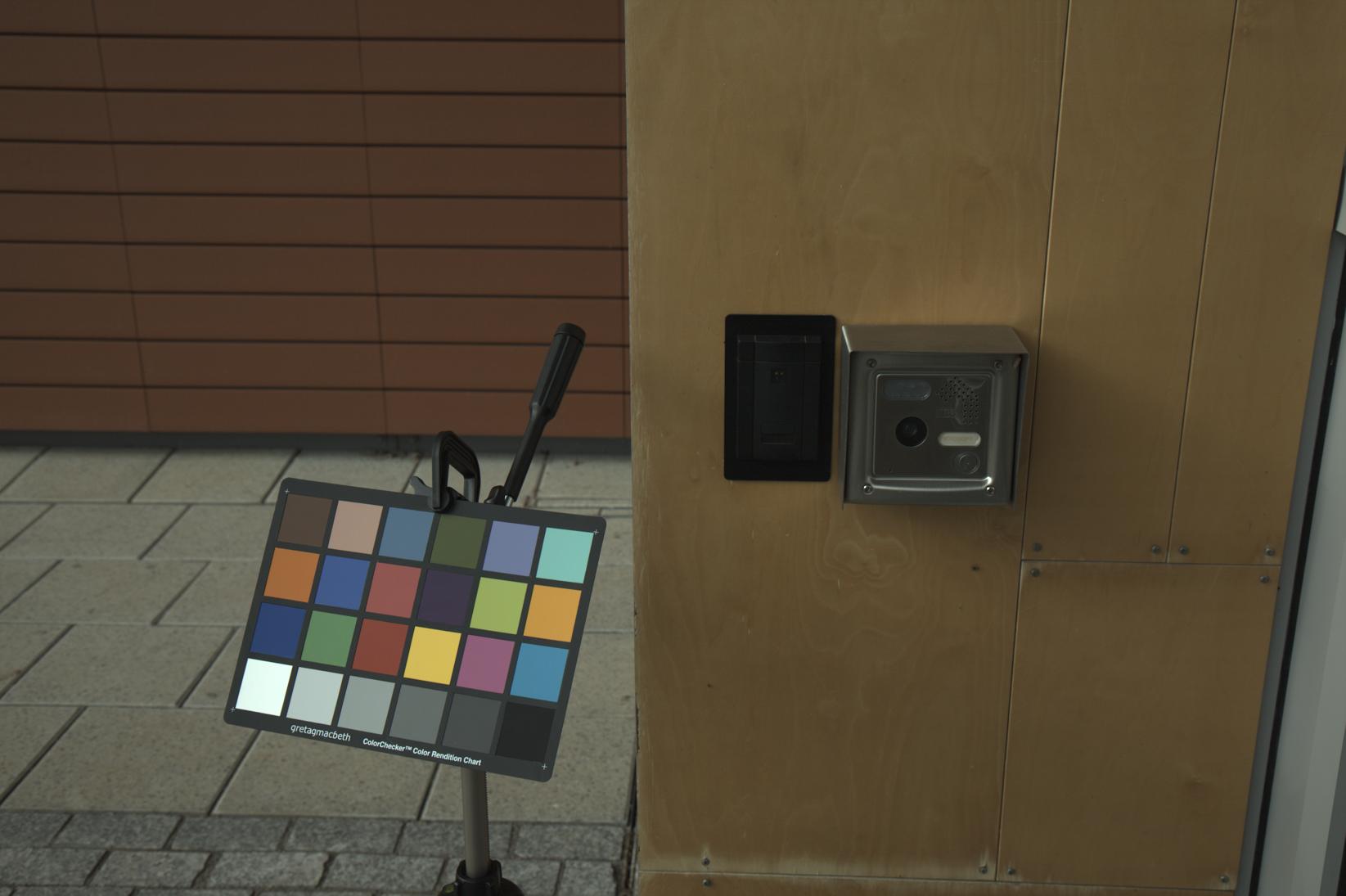} & \includegraphics[width=0.40\columnwidth]{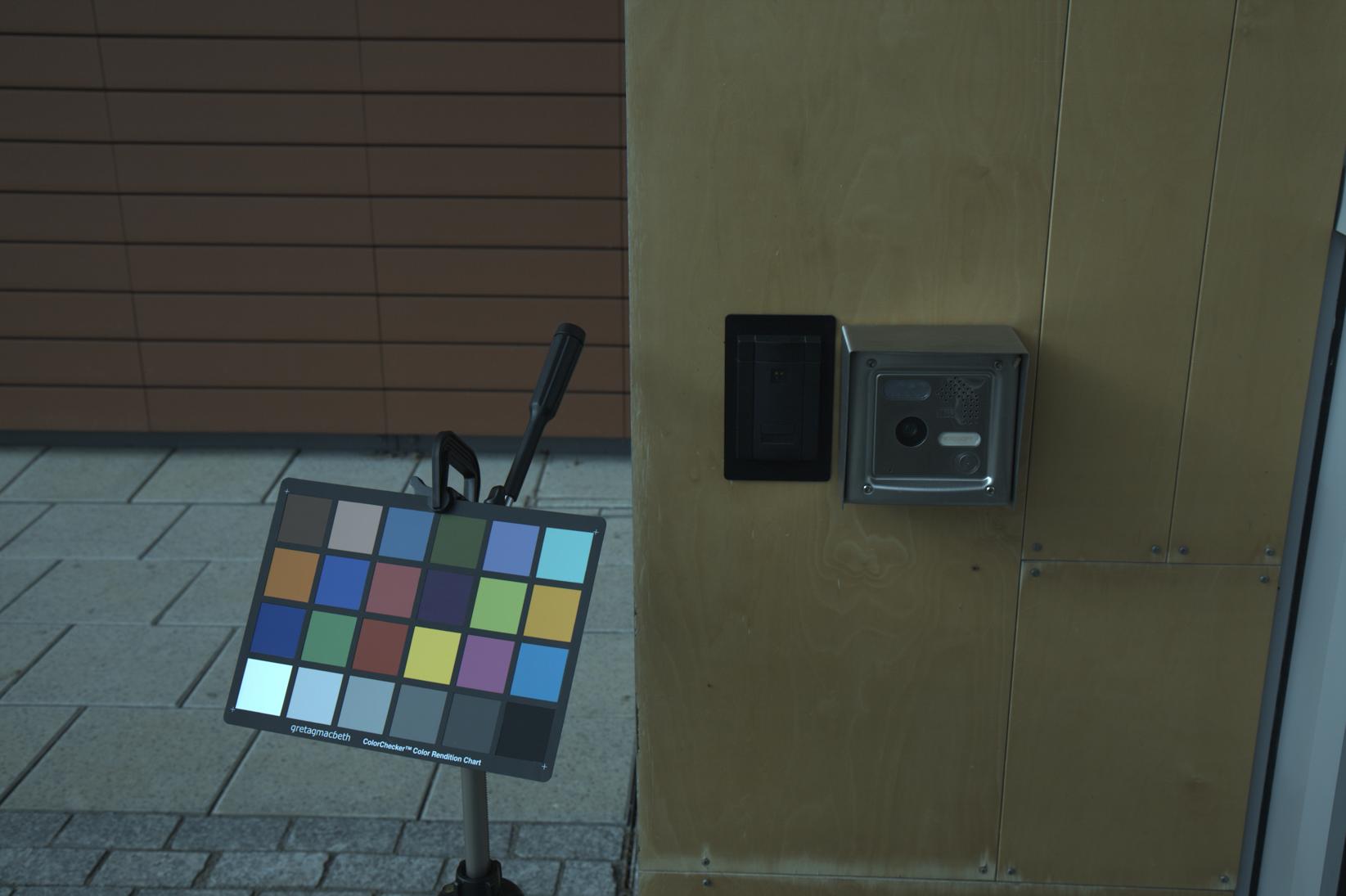} & \includegraphics[width=0.40\columnwidth]{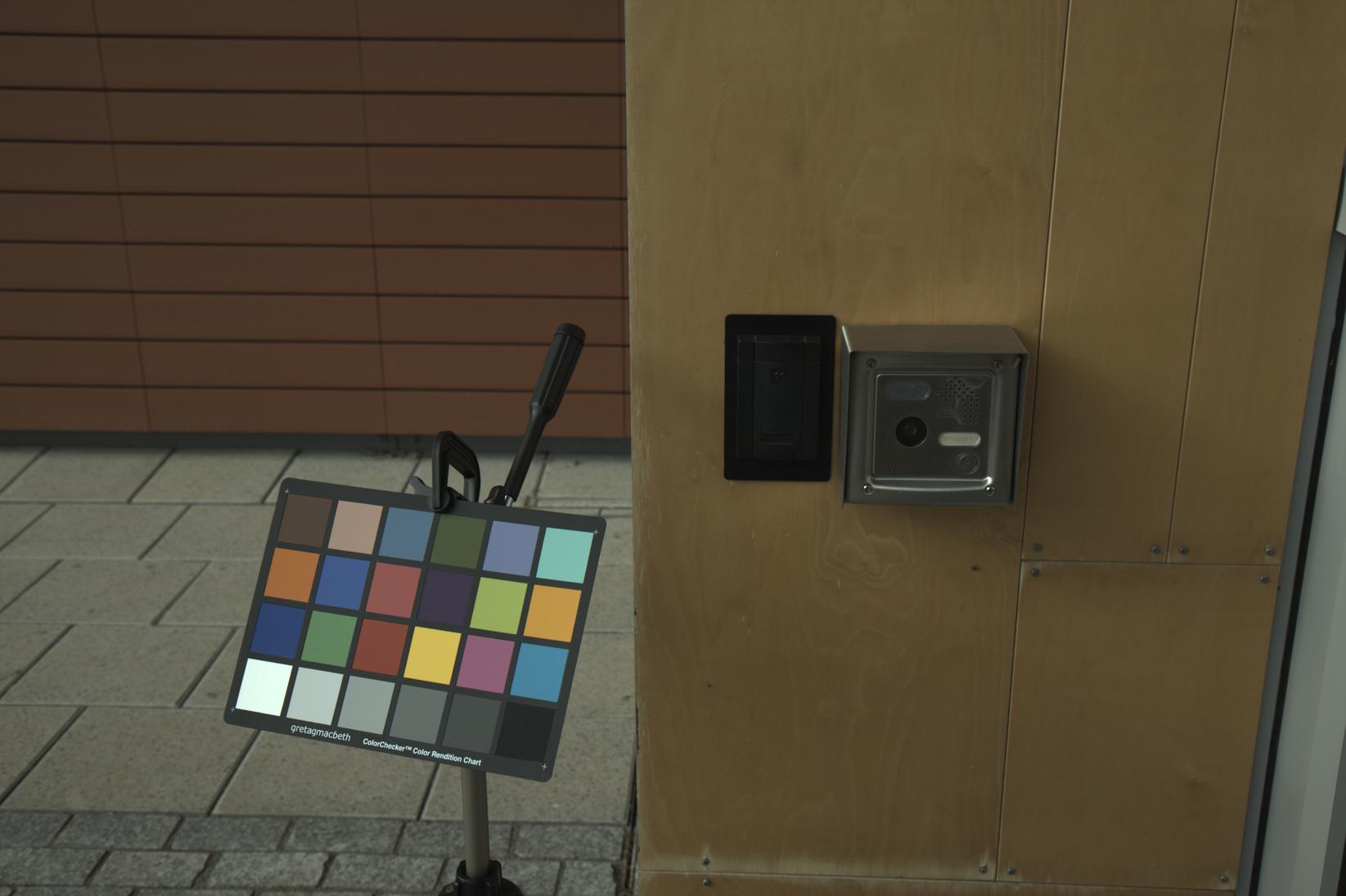} \\
		Original & groundtruth (0$^\circ$) & CNN fine-tuned (10.87$^\circ$) & GM (0.52$^\circ$)\\

		\includegraphics[width=0.40\columnwidth]{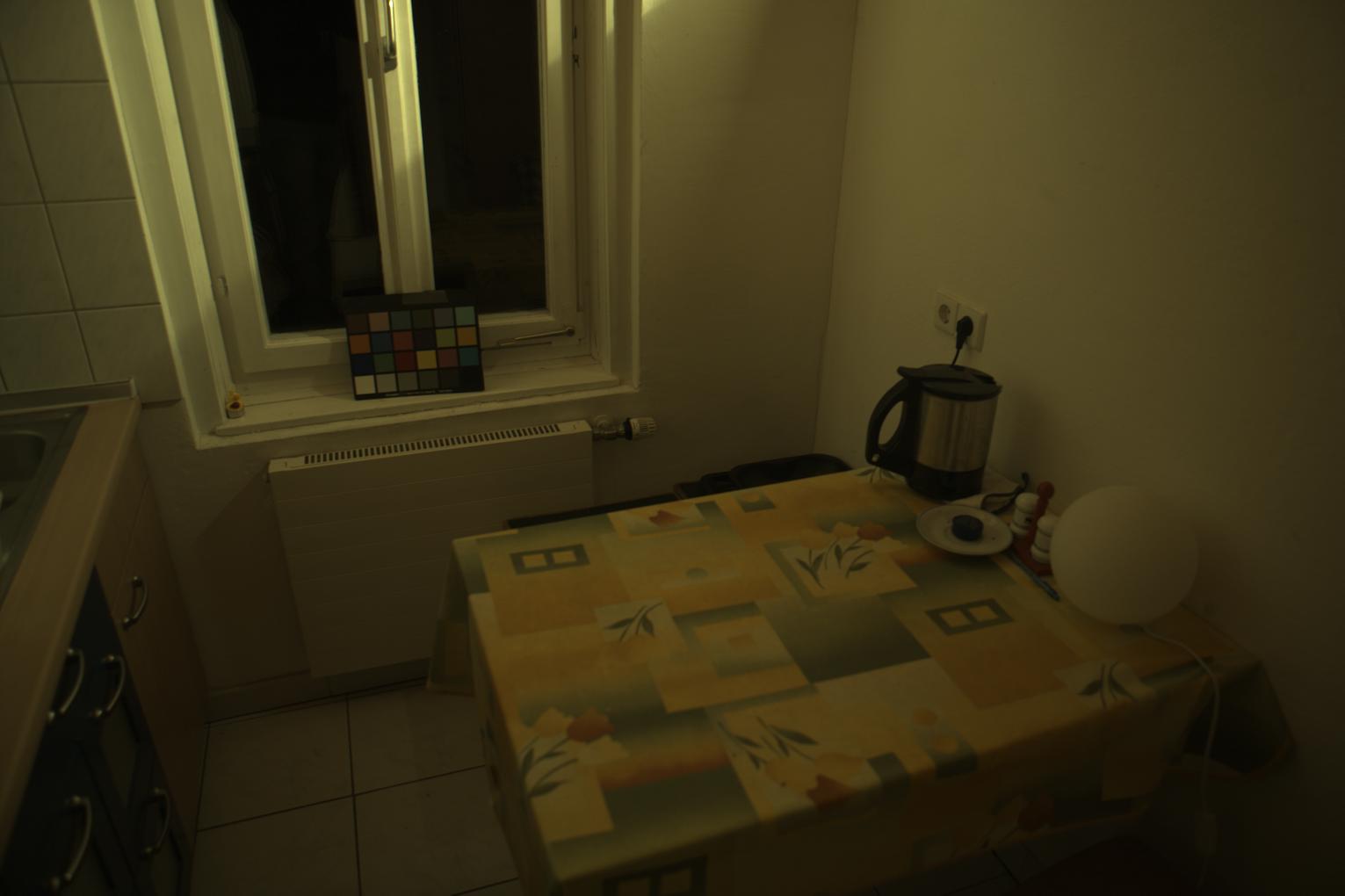} & \includegraphics[width=0.40\columnwidth]{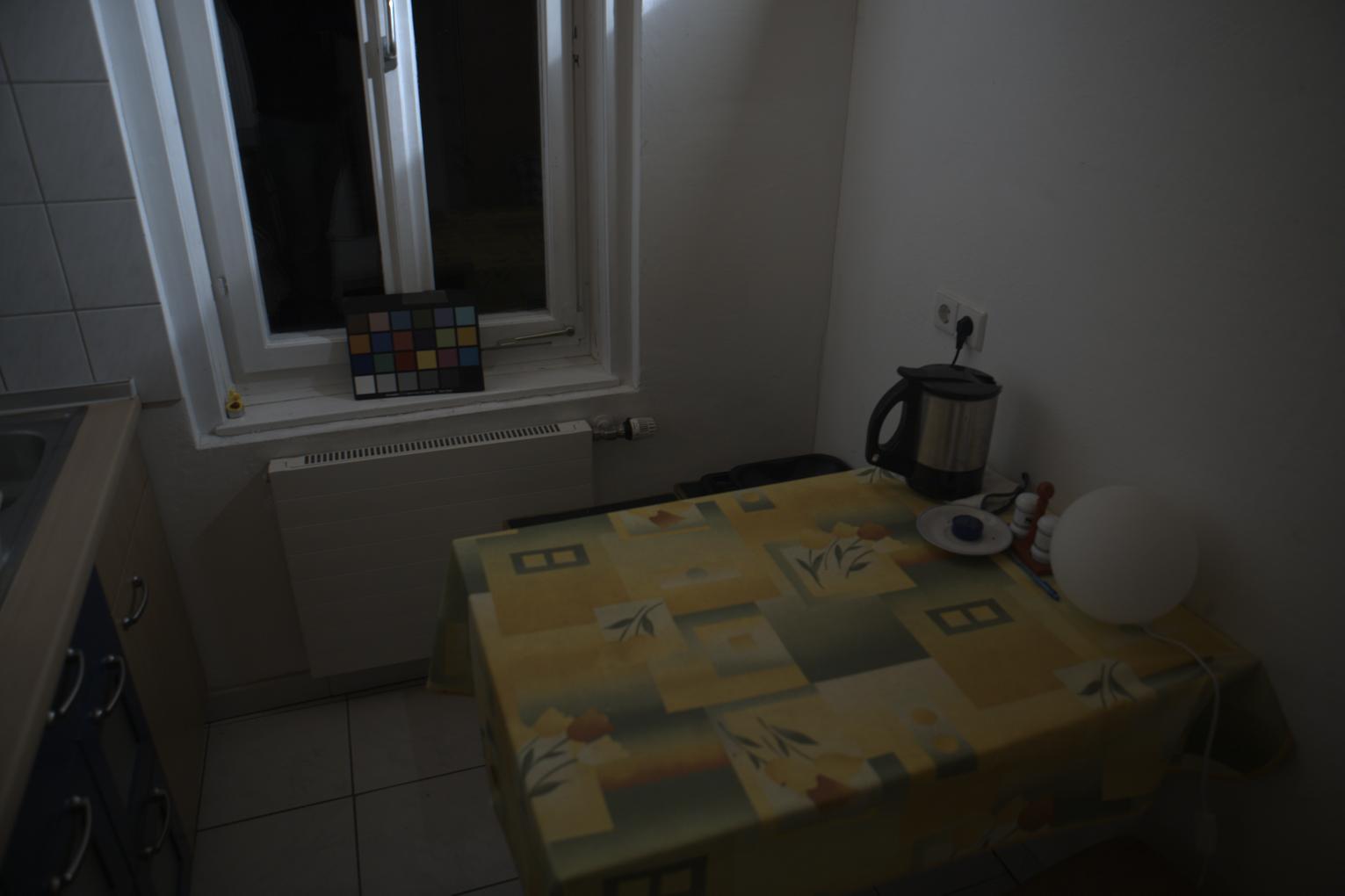} & \includegraphics[width=0.40\columnwidth]{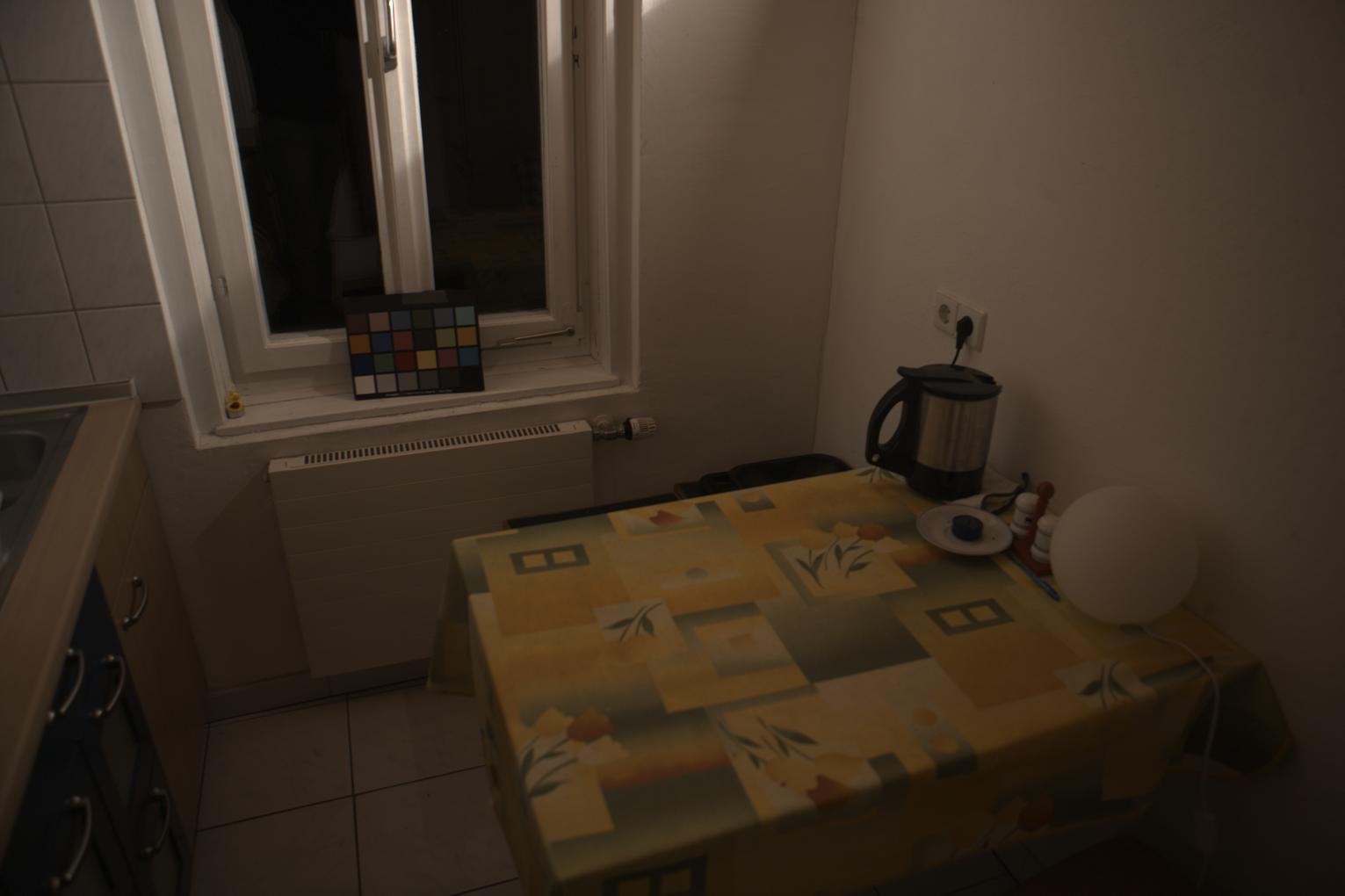} & \includegraphics[width=0.40\columnwidth]{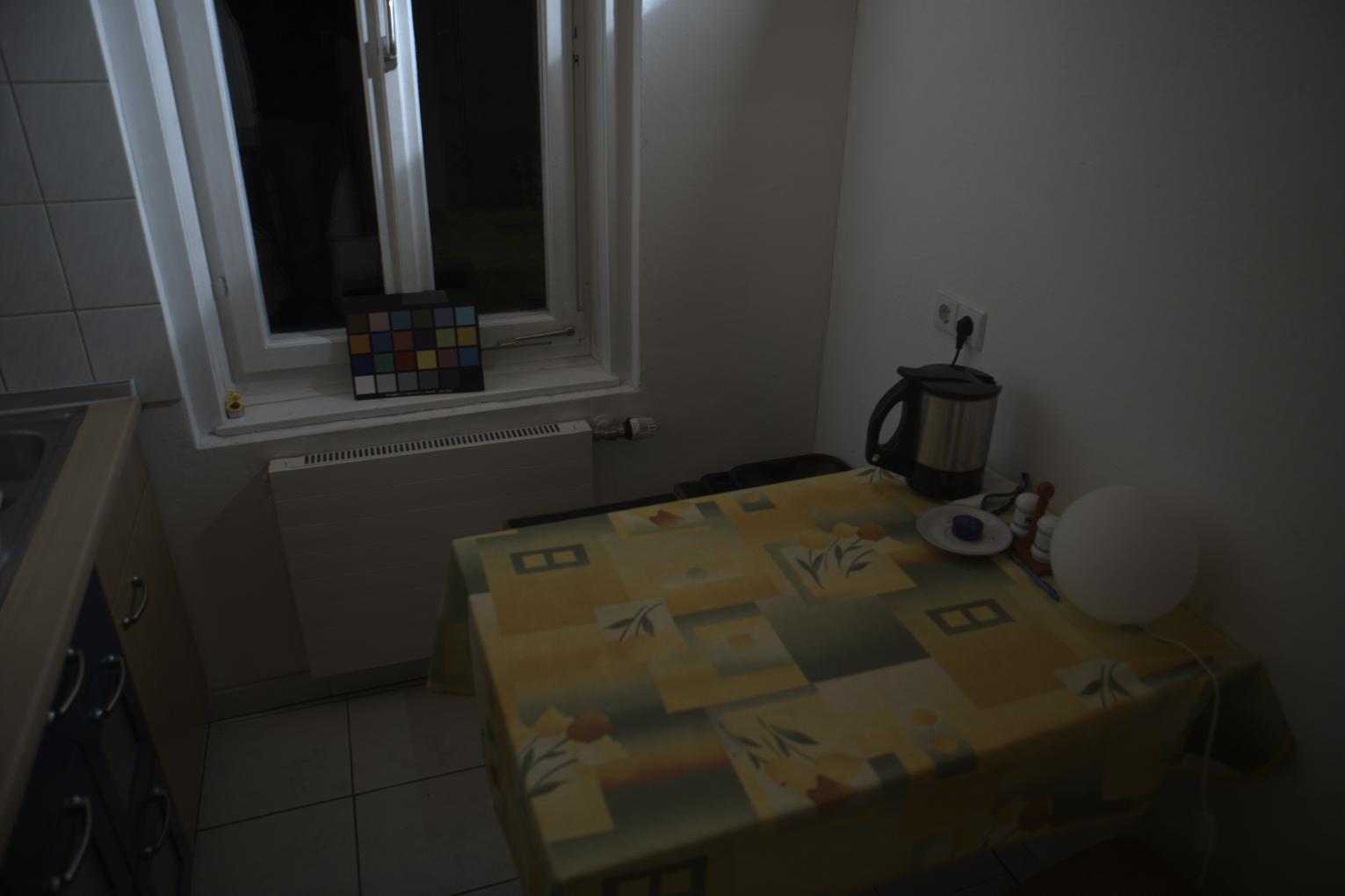} \\
		Original & groundtruth (0$^\circ$) & CNN fine-tuned (10.85$^\circ$) & HLVI BU (0.42$^\circ$)\\

		\includegraphics[height=0.40\columnwidth,angle=90]{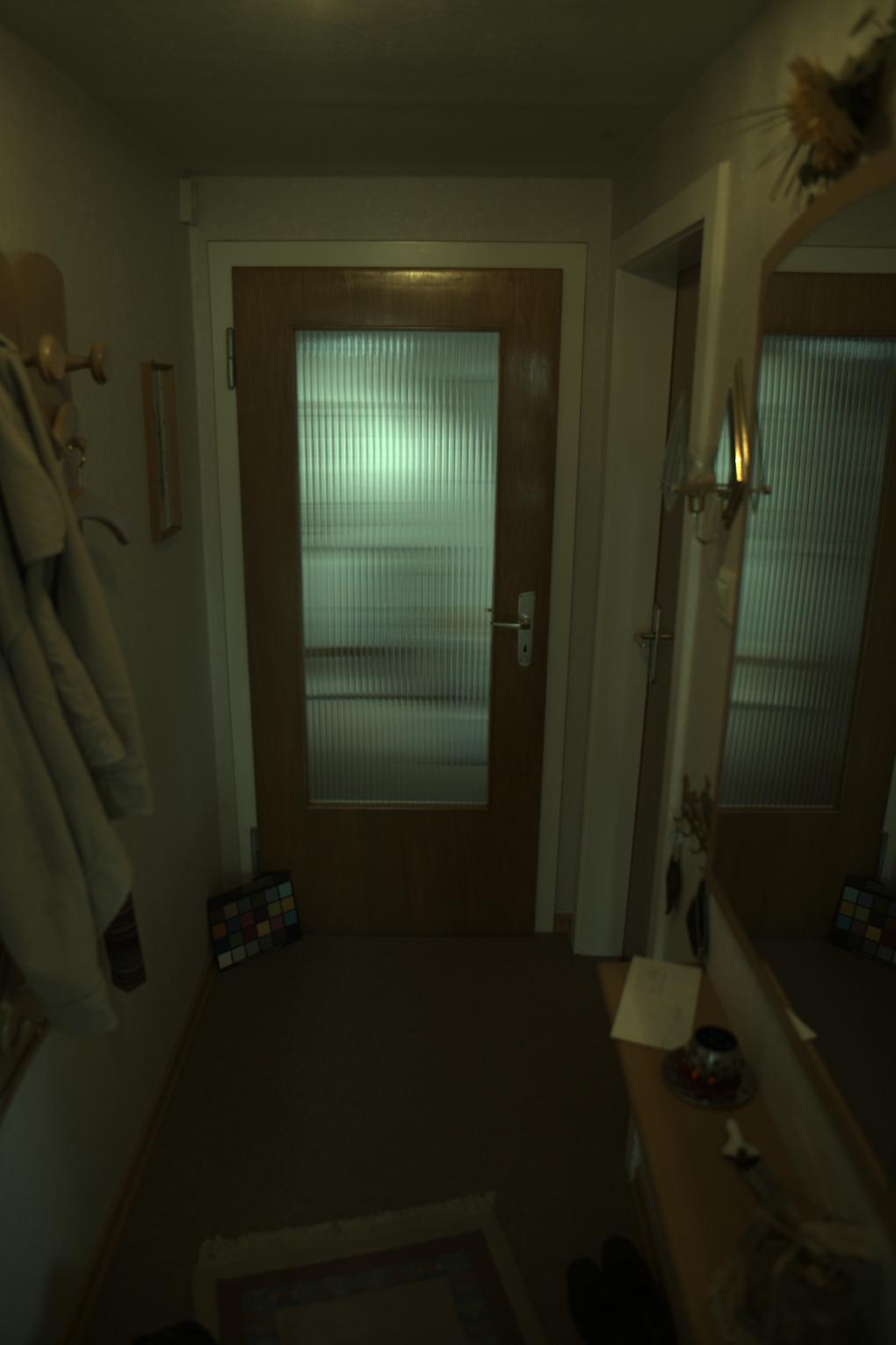} & \includegraphics[height=0.40\columnwidth,angle=90]{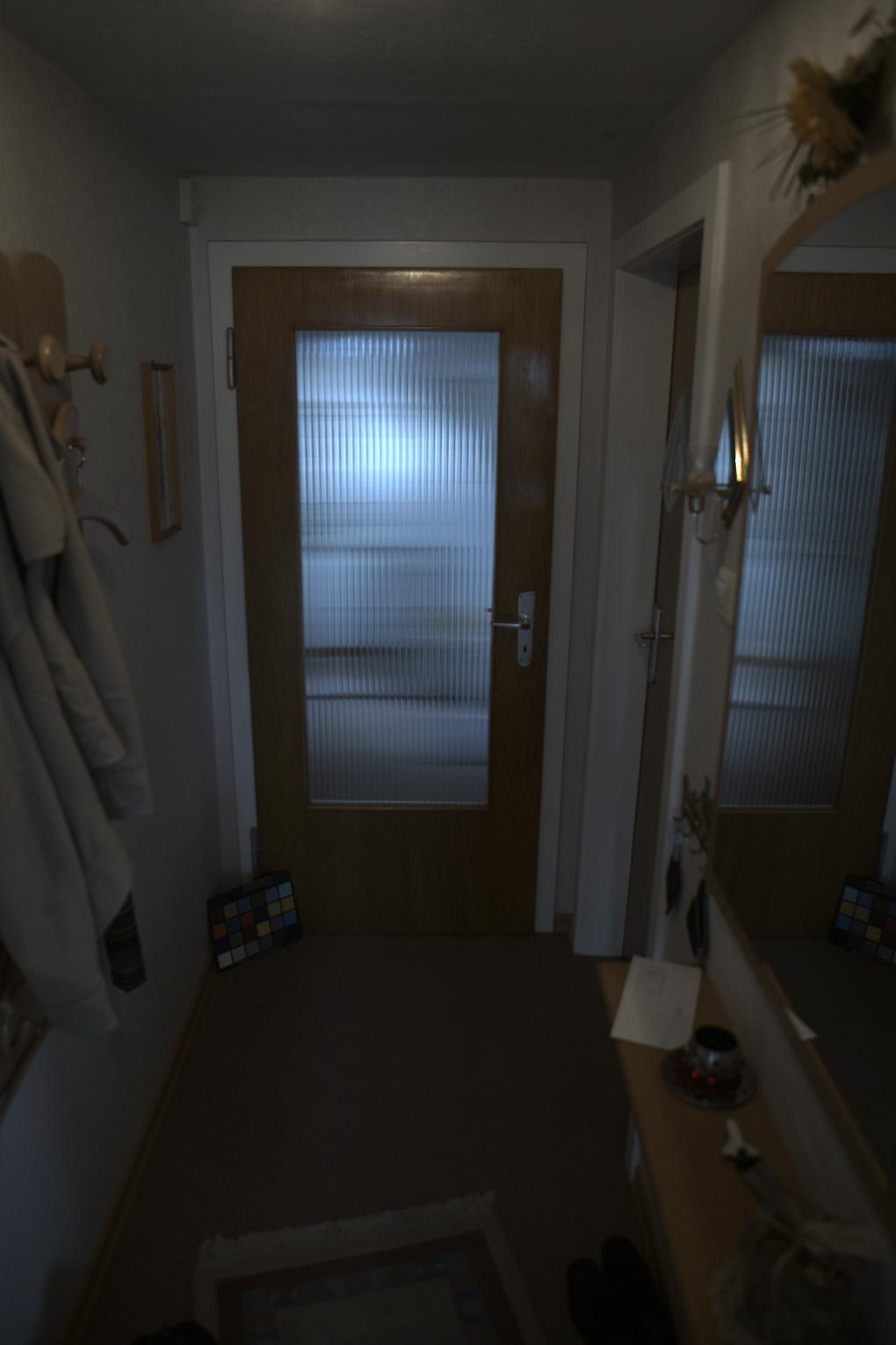} & \includegraphics[height=0.40\columnwidth,angle=90]{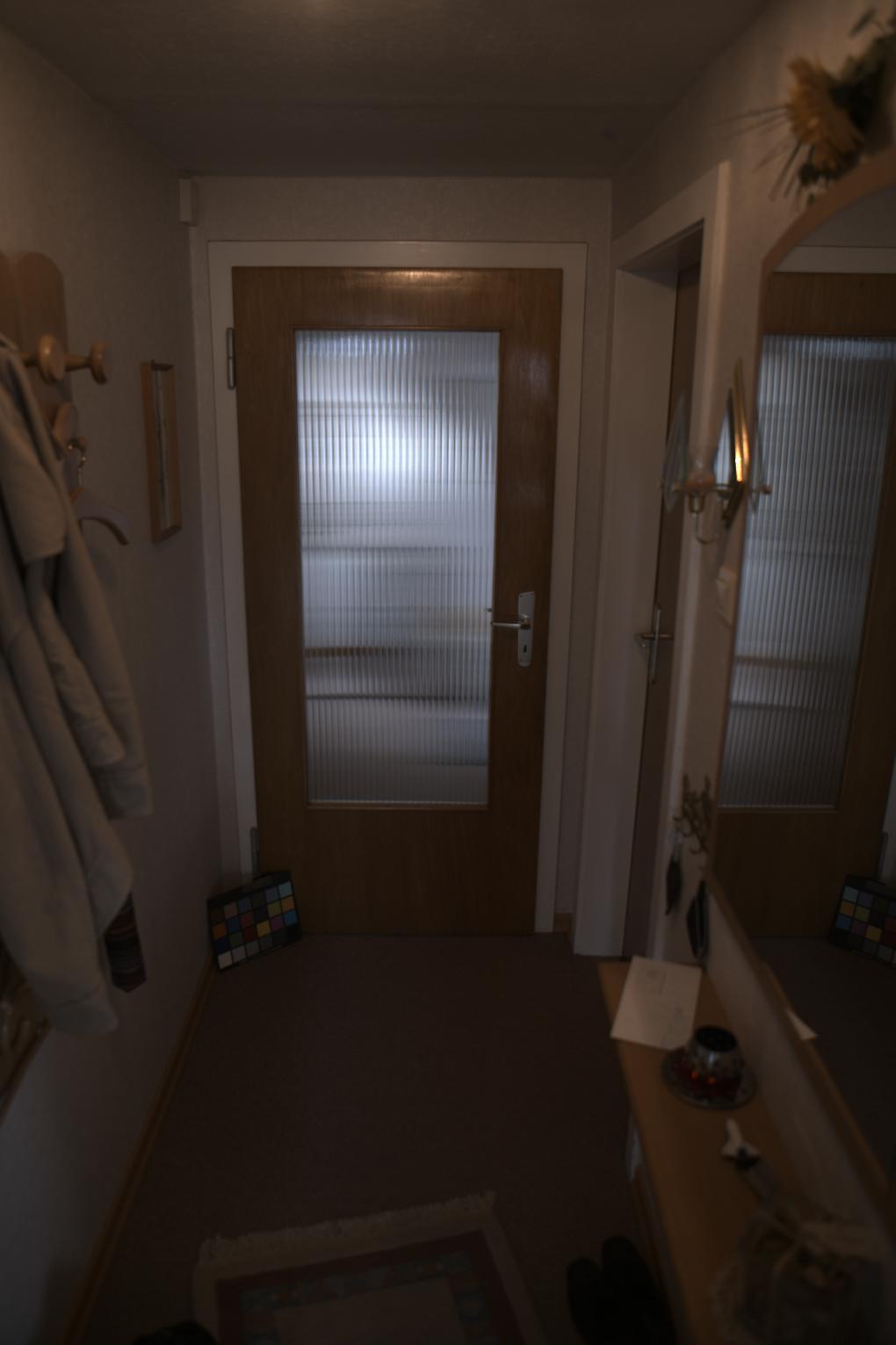} & \includegraphics[height=0.40\columnwidth,angle=90]{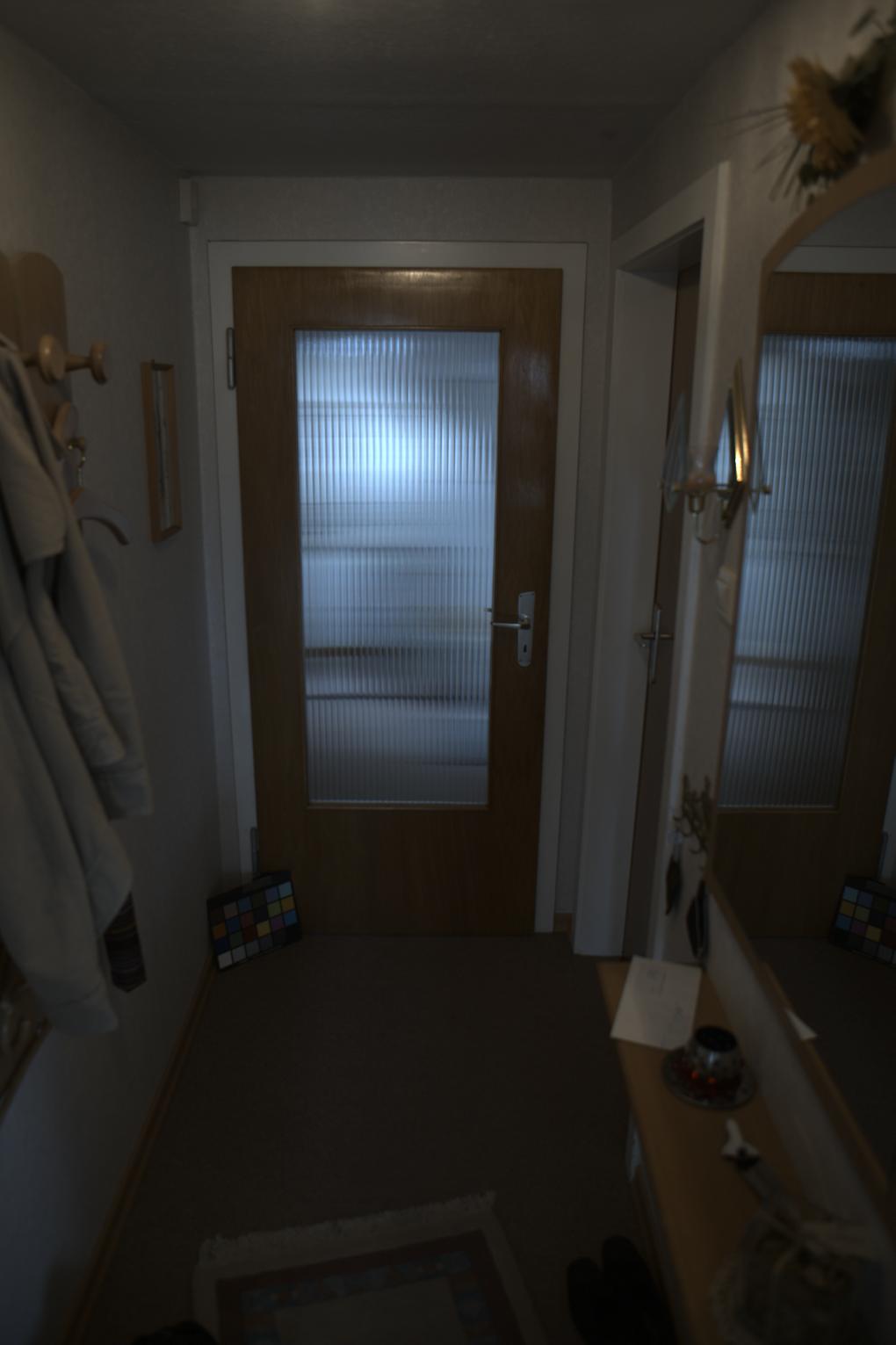}\\
		Original & groundtruth (0$^\circ$) & CNN fine-tuned (9.93$^\circ$) & GW (1.35$^\circ$)\\

		\includegraphics[height=0.40\columnwidth,angle=90]{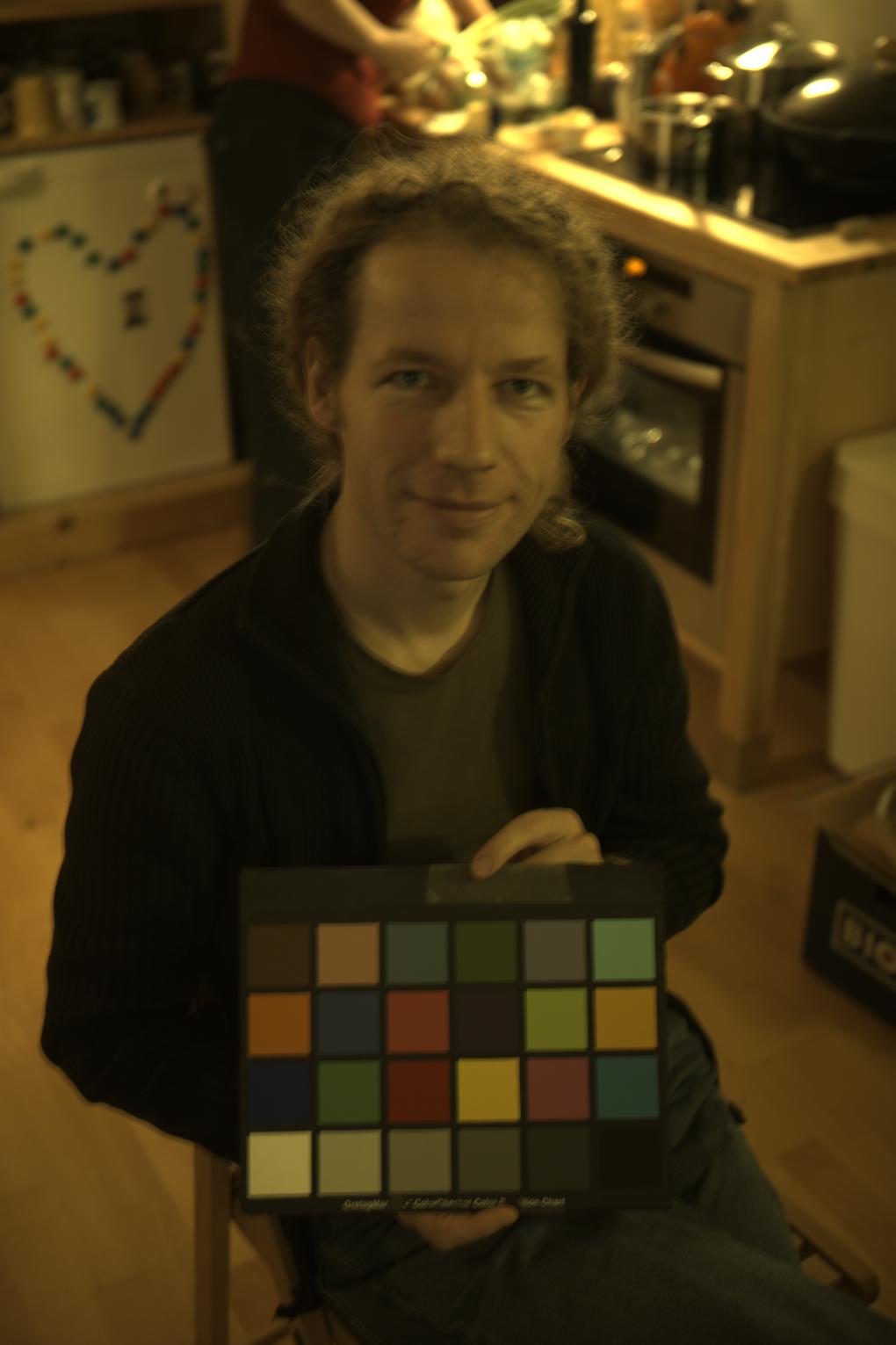} & \includegraphics[height=0.40\columnwidth,angle=90]{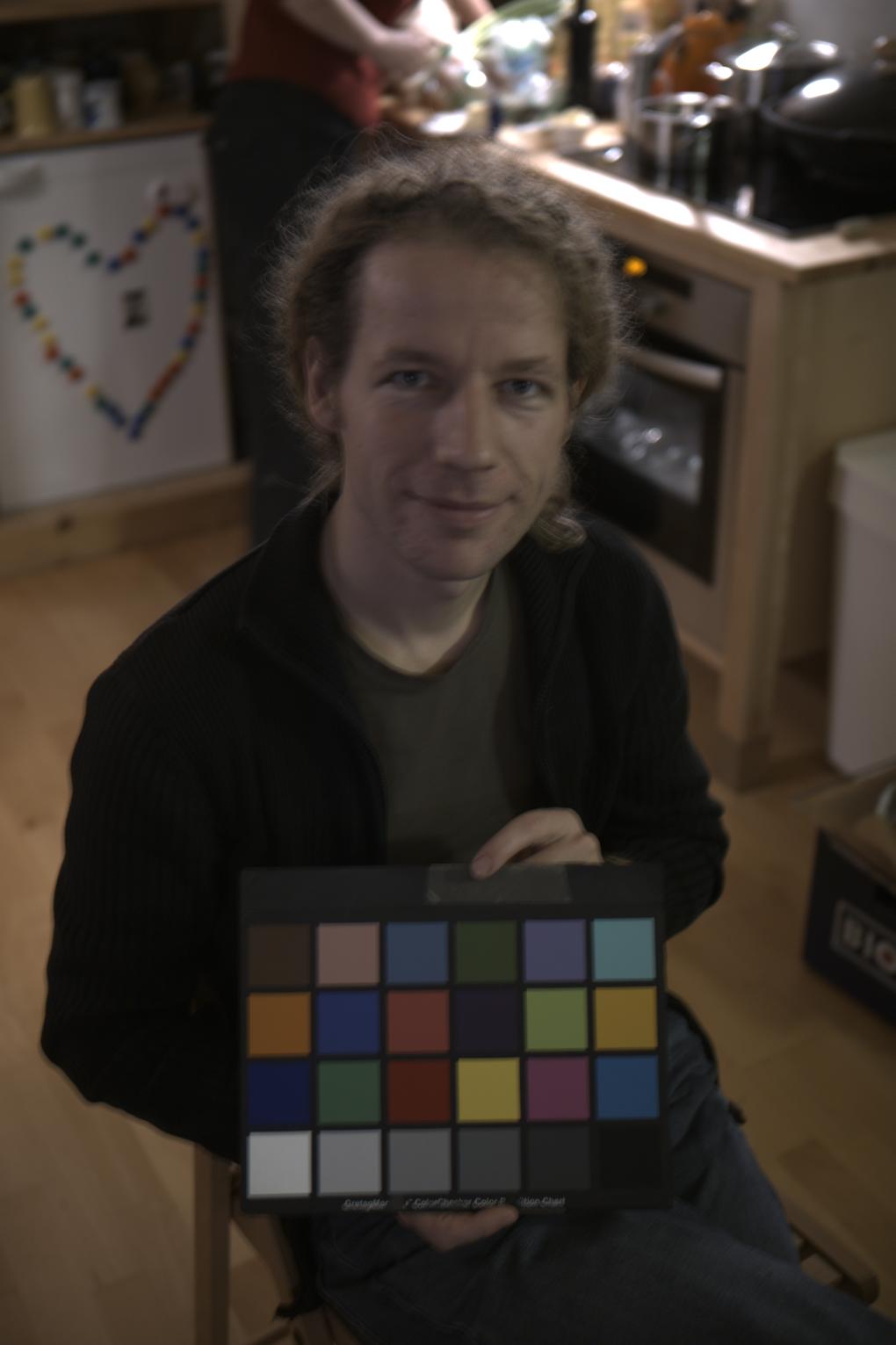} & \includegraphics[height=0.40\columnwidth,angle=90]{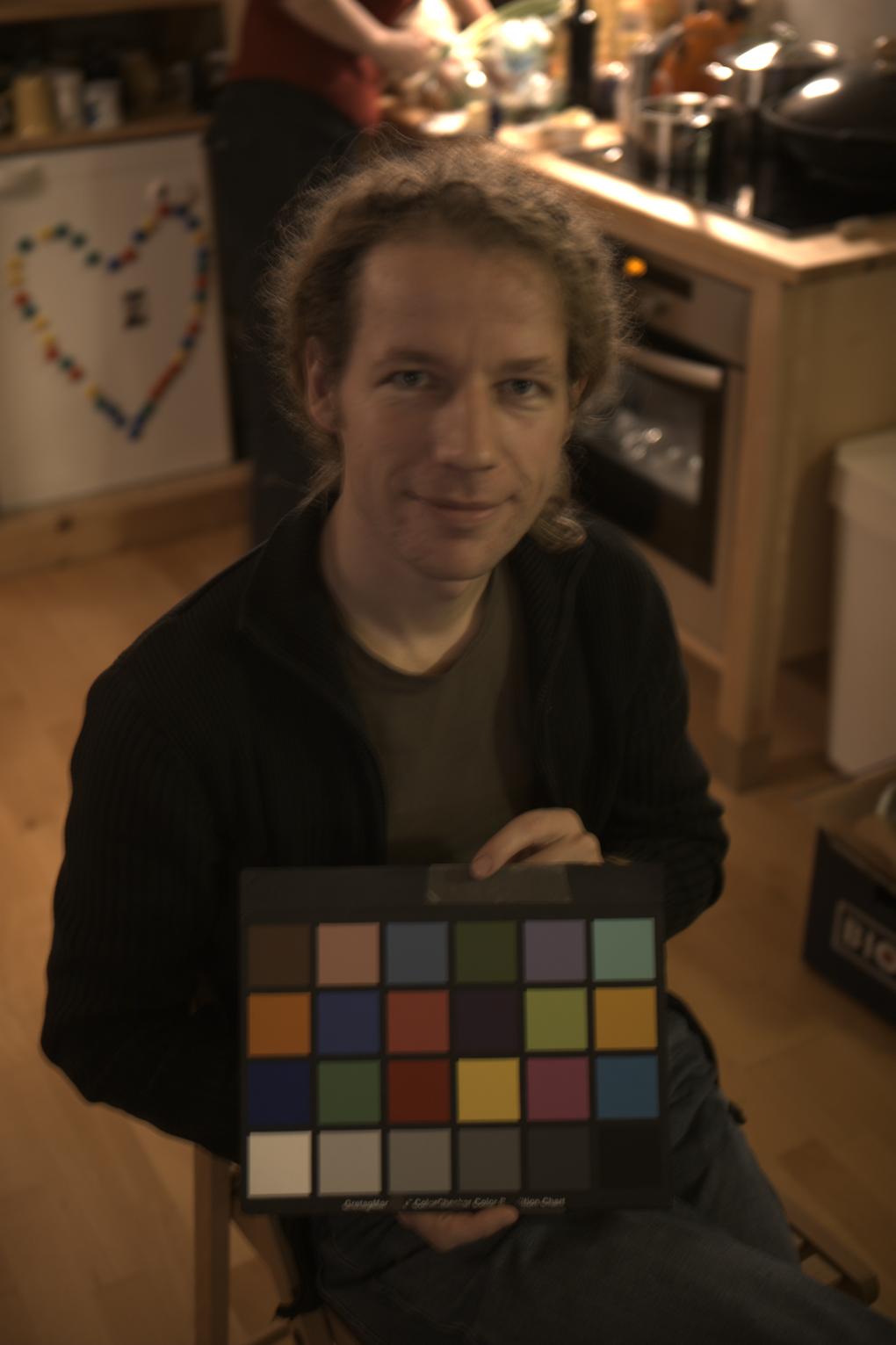} & \includegraphics[height=0.40\columnwidth,angle=90]{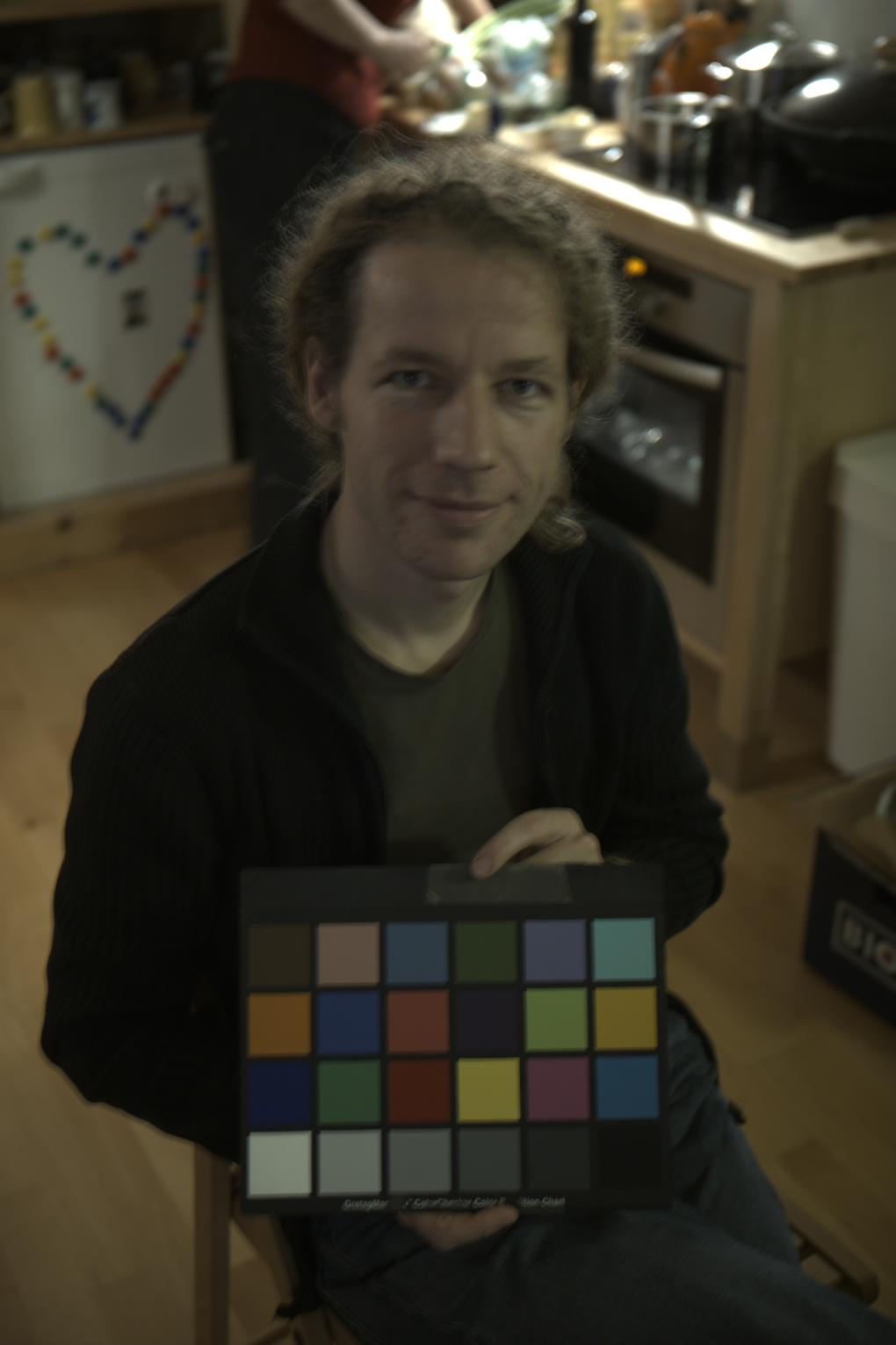}\\
		Original & groundtruth (0$^\circ$) & CNN fine-tuned (9.85$^\circ$) & NIS (3.32$^\circ$)\\

	\end{tabular}
	\caption{Examples of images on which the fine-tuned CNN makes the largest estimation errors. Left to right: original RAW image, correction with the groundtruth illuminant, correction with the CNN estimate, and correction with the algorithm in the state-of-the-art making the best estimate on the given image.}
	\label{fig:worstErrors}
\end{figure*}

Figure \ref{fig:worstErrors} reports some examples of images on which the fine-tuned CNN makes the largest estimation errors. From left to right we report the original RAW image, the image corrected with the groundtruth illuminant, the image corrected with the CNN estimate, and the image corrected with the algorithm in the state-of-the-art making the best estimate on that image.
Once we have an estimate of the global illuminant color $\mathbf{I}$, each pixel in the image is color corrected using the von Kries model \cite{von1902chromatic}, i.e.: $\boldsymbol{\rho}_{out}(x,y)=diag(\mathbf{I}^{-1})\boldsymbol{\rho}_{in}(x,y)$. 

\begin{table*}[!ht]
\caption{Angular error statistics obtained by the state-of-the-art algorithms considered on the RAW dataset.}
\label{tab:errori}
\centering
\resizebox{1.25\columnwidth}{!} {
\begin{tabular}{lrrrrrr}
  \toprule
	Algorithm & Min & 10$^{th}$prc &  Med &  Avg & 90$^{th}$prc &  Max  \\
  \midrule
DN    												& 3.72 &  10.38 &  13.55 &  13.62 &  16.45 &  27.37  \\
GW    												& 0.18 &   1.88 &   6.30 &   6.27 &  10.12 &  24.84  \\
WP    												& 0.08 &   1.38 &   5.61 &   7.46 &  15.68 &  40.59  \\
SoG   												& 0.18 &   1.04 &   4.04 &   4.85 &   9.71 &  19.93  \\
gGW   												& \bf{0.03} &   0.82 &   3.45 &   4.60 &   9.68 &  22.21  \\
GE1  												  & 0.16 &   1.82 &   4.55 &   5.21 &   9.78 &  19.69  \\
GE2   												& 0.26 &   2.06 &   4.43 &   5.01 &   \bf{8.93} &  \bf{16.87}  \\
GM    												& 0.05 &   \bf{0.40} &   \bf{2.28} &   \bf{4.10} &  11.08 &  23.18  \\
\midrule
SVR							& 0.66 & 3.36 & 6.67 & 7.99 & 14.61 & 26.08 \\
{BAY} 					& 0.10  &  1.17 &   3.44 &   4.70 &  10.21 &  24.47 \\
{NIS}      		& 0.08  &  0.93 &   3.13 &   4.09 &   8.57 &  26.20 \\
{HLVI BU} 			& 0.06  &  0.75 &   2.54 &   3.30 &   6.59 &  17.51 \\
{HLVI TD} 			& 0.11  &  0.85 &   2.63 &   3.65 &   7.53 &  25.24 \\
{HLVI BU\&TD}  & 0.13  &  0.77 &   2.47 &   3.38 &   6.97 &  25.24 \\
{SS ML}        & 0.06 &   0.85 &   2.93 &   3.55 &   7.23 &  15.25 \\
{SS GP}   			& 0.07 &   0.82 &   2.90 &   3.47 &   7.00 &  \bf{14.80} \\
{AAS}				& \bf{0.03} &   0.77 &   3.16 &   4.18 &   9.15 &  22.21 \\
{AAC}				& 0.05 &   0.90 &   2.90 &   3.74 &   7.93 &  14.98 \\
EB 		& 0.14 & 0.73   & 2.24   & \bf{2.77}   &  \bf{5.52}  &  19.44 \\
FB+GM    					& 0.05  &  \bf{0.40} &   \bf{2.01} &   3.67 &   9.50 &  23.18  \\
{FB+SS GP}   			 & 0.08 &   0.75 &  2.57 & 3.18 & 6.67 & \bf{14.80} \\

\midrule
AlexNet+SVR				 & 0.12  & 0.98  & 3.09  & 4.74 & 11.18 & 29.15\\ 
CNN per patch				 &\bf{0.00}  & 0.99  & 2.69  & 3.67 &  7.79 & 30.93 \\
CNN	average-pooling					 & 0.04  & 0.99  & 2.44  & 3.18 &  6.37 & 14.84 \\ 
CNN median-pooling					 & 0.06  & 0.97  & 2.32  & 3.07 &  6.15 & 19.04 \\
CNN fine-tuned	 			 & 0.06  & \bf{0.69}  & \bf{1.98}  & \bf{2.63} &  \bf{5.54} & \bf{14.77} \\ 
\bottomrule
\end{tabular}
}
\end{table*} 


\subsection{Effects of parameters}
\begin{figure*}
	\centering
	\setlength{\tabcolsep}{-3pt}
	\resizebox{\textwidth}{!}{
	\begin{tabular}{ccc}
		\includegraphics[width=0.63\columnwidth]{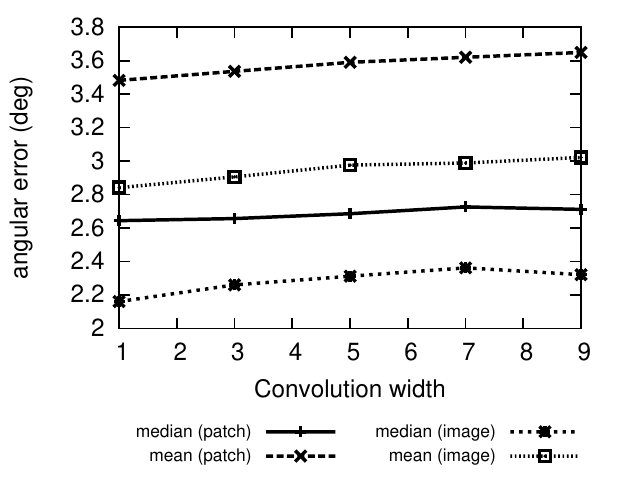} &
		\includegraphics[width=0.63\columnwidth]{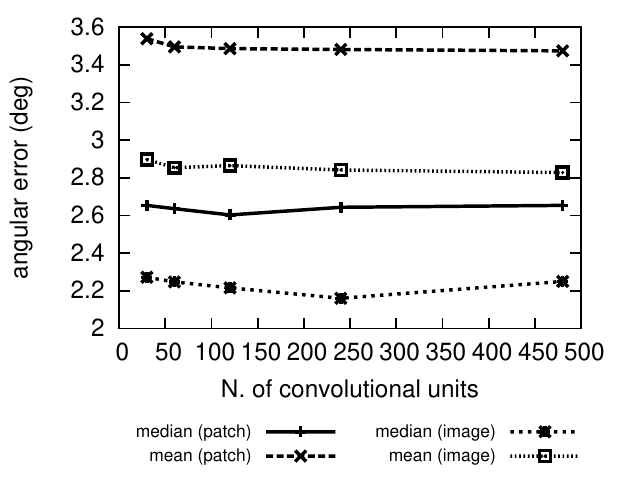} &
		\includegraphics[width=0.63\columnwidth]{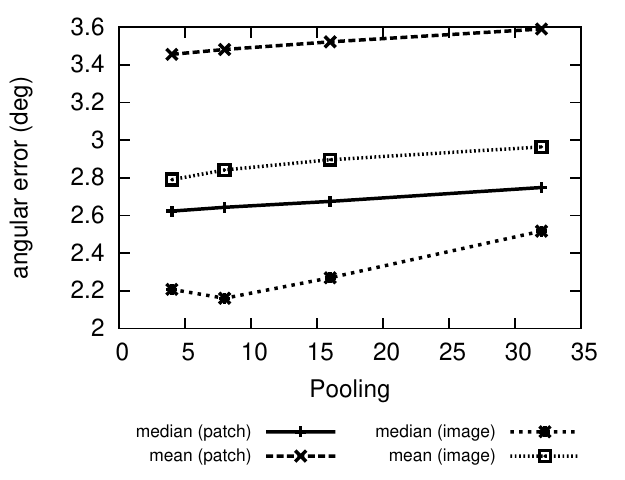} \\
		a) & b) & c) \\
		\includegraphics[width=0.63\columnwidth]{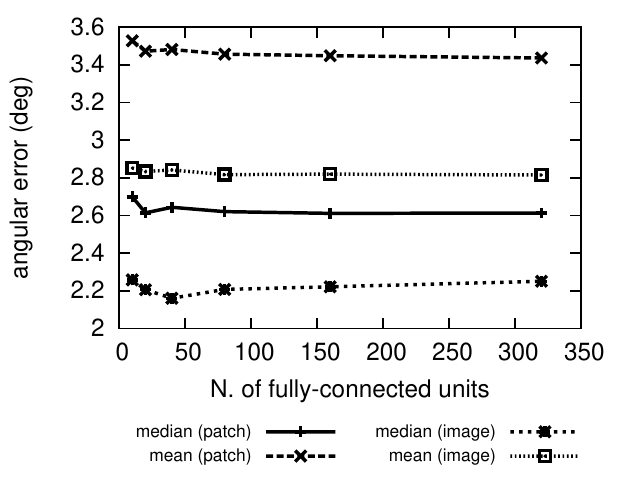} &
		\includegraphics[width=0.63\columnwidth]{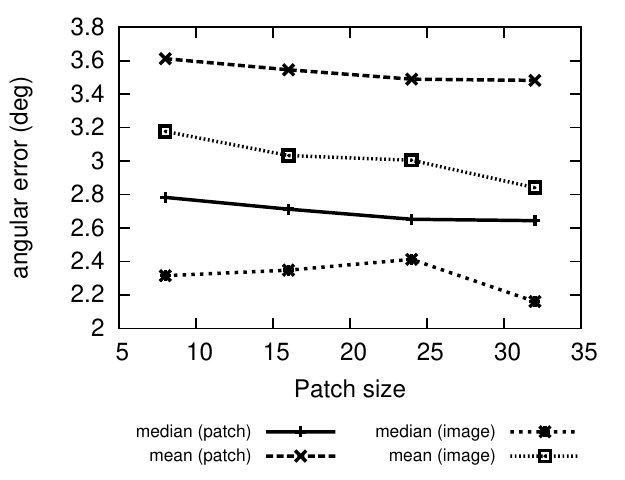} &
		\\
		d) & e) & \\
	\end{tabular}
	}
	\caption{Effects of the parameters on the CNN performance. Angular error with respect to varying convolution kernel width (a), number of convolutional kernels (b), pooling size (c), number of fully connected units (d) and input patch size (e). Each point corresponds to the best performance that can be obtained by fixing a single parameter at the value indicated by trying all the combinations for the other ones.}
	\label{fig:parametri}
\end{figure*}

Several parameters are involved in the CNN design. In
this section, we examine how these parameters affect the
network performance. 
The network architecture has been chosen by starting from a 7-layer deep CNN similar to \cite{krizhevsky2012imagenet} and removing layers until no further improvement in performance was possible.
In Figure \ref{fig:parametri} we report how the different parameters affect the illuminant estimation performance. 
Each point on the graphs represents the best result that can be obtained by fixing a parameter at the value indicated and searching over all the possible combinations of the other ones.

{\bf{Kernel size}} Figure \ref{fig:parametri}.a shows how the performance
varies with the width of the convolution kernels. 
We can see from Figure \ref{fig:parametri}.a that the estimation error decreases by decreasing the kernel size. 
At first this could be surprising, since in different domains larger kernels are preferred. However, it is not the first time that such small kernels are used, see \cite{szegedy2014going}. From the color constancy point of view, this choice of kernel size confirms the finding of Cheng at al. \cite{cheng2014illuminant}, where they showed that spatial information does not provide any additional information that cannot be obtained directly from the color distributions. 

{\bf{Number of kernels}} Figure \ref{fig:parametri}.b shows how the performance
varies with respect to the number of convolution kernels. It is
possible to see that the CNN tends to prefer an intermediate number of kernels.

{\bf{Pooling size}} Figure \ref{fig:parametri}.c shows how the performance
varies with respect to the number of convolution kernels. It is
possible to see that the CNN tends to prefer an intermediate pooling size.

{\bf{Number of fully connected units}} Figure \ref{fig:parametri}.d shows how the performance
varies with respect to the number of fully connected units. The plot shows that better performance can be reached with a number of fully connected units around 40.

{\bf{Patch size}} Since in our experiment the illuminant is estimated as the median illuminant of all patches sampled,
we examine how the patch size affects performance.
For every patch size, the same number of patches is randomly extracted from each image being sure that none of them contained the reference color chart.  
{Figure \ref{fig:parametri}.e shows the change of performance with respect to patch size. From the plot we see that a larger patch size results in better performance. 
}

\subsection{Local illuminant estimation}
Our CNN predicts the illumination on small image patches, so it can be easily used to predict  local illuminants as well as giving a global illuminant estimate for the entire image.
Given the performance of the per patch error in Table \ref{tab:errori} we expect our CNN to perform well even on local estimation. We perform here a preliminary test by using our learned CNN as-is on a dataset of synthetic images: the images are taken from the previous RAW dataset \cite{Shi} and on half of each image we manually changed the illuminant, resulting in two illuminants for each image.

Among the algorithms in the state-of-the-art able to deal with non-uniform illumination, e.g. \cite{retinex,provenzi2008spatially,LSAC,Bleier,joze2014exemplar,bianco2014adaptive} we report as comparison the results of the Multiple Light Sources (MLS) \cite{MLS} using White Point (WP) and Gray World (GW) algorithms, grid based sampling, in the clustering version setting the number of clusters equal to the number of lights in the scene, i.e. two.

\begin{figure*}
	\centering
	\begin{tabular}{cccc}
\includegraphics[width=0.45\columnwidth]{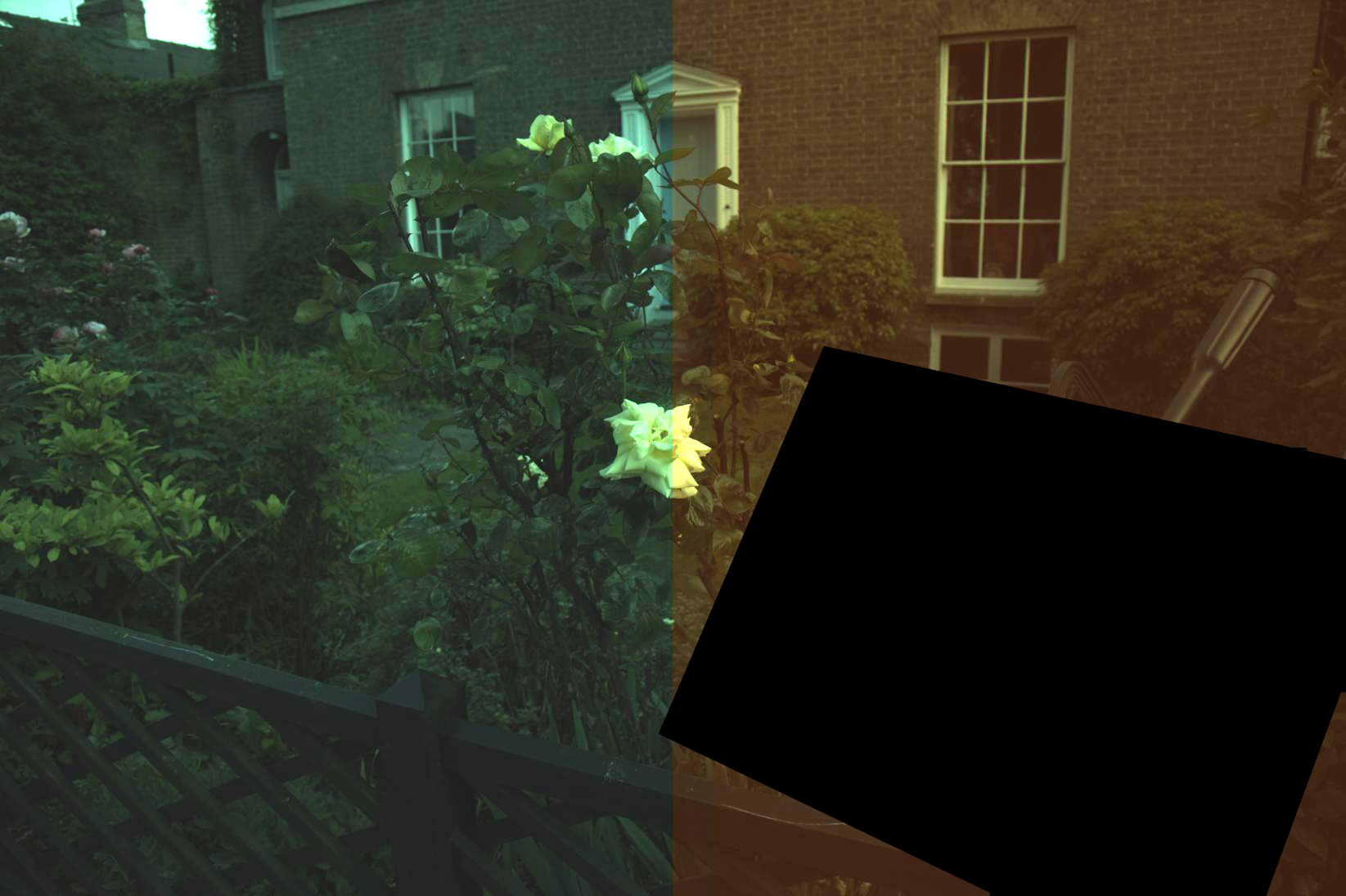} 			 & \includegraphics[width=0.45\columnwidth]{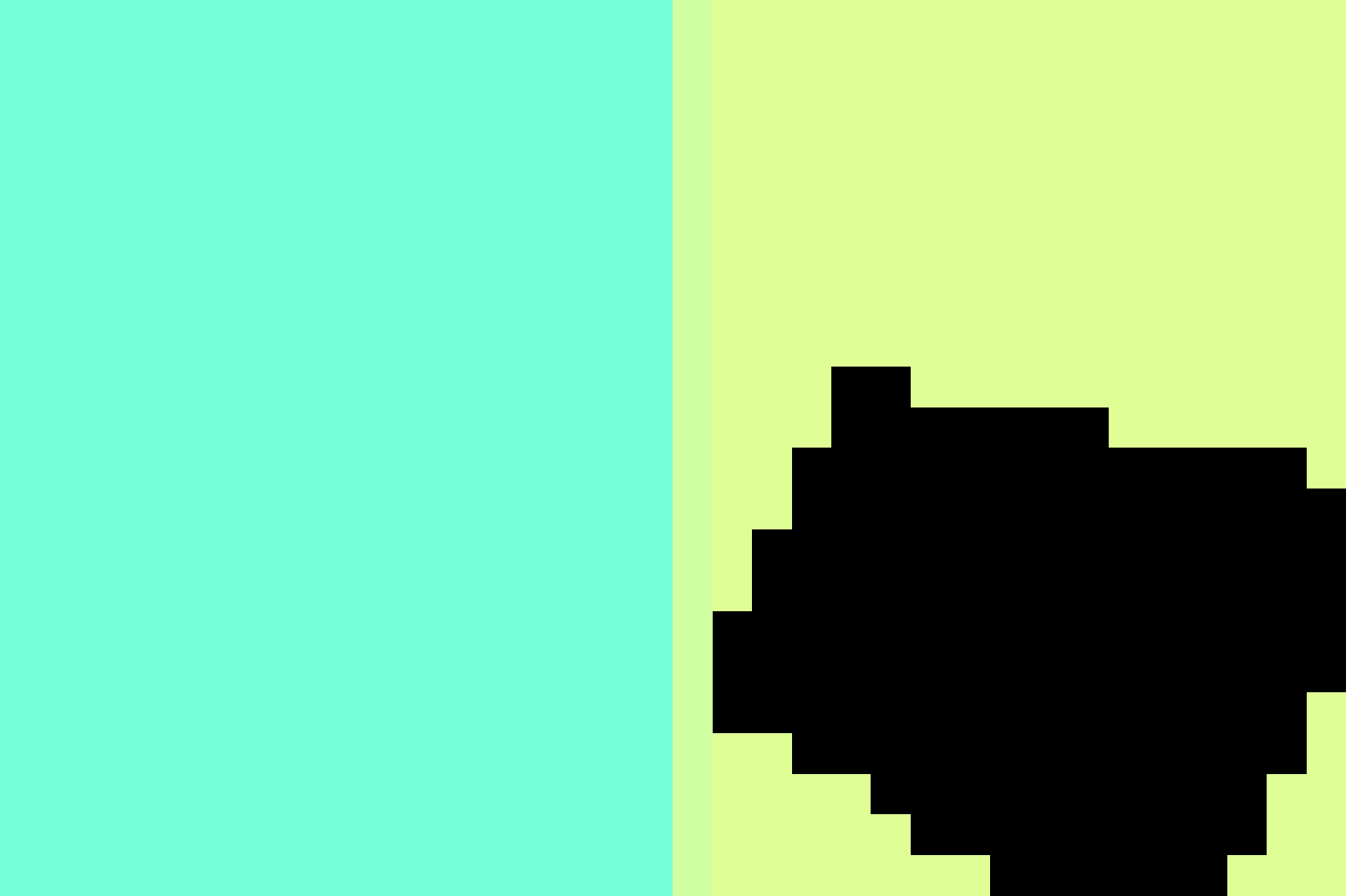} & 
\includegraphics[width=0.45\columnwidth]{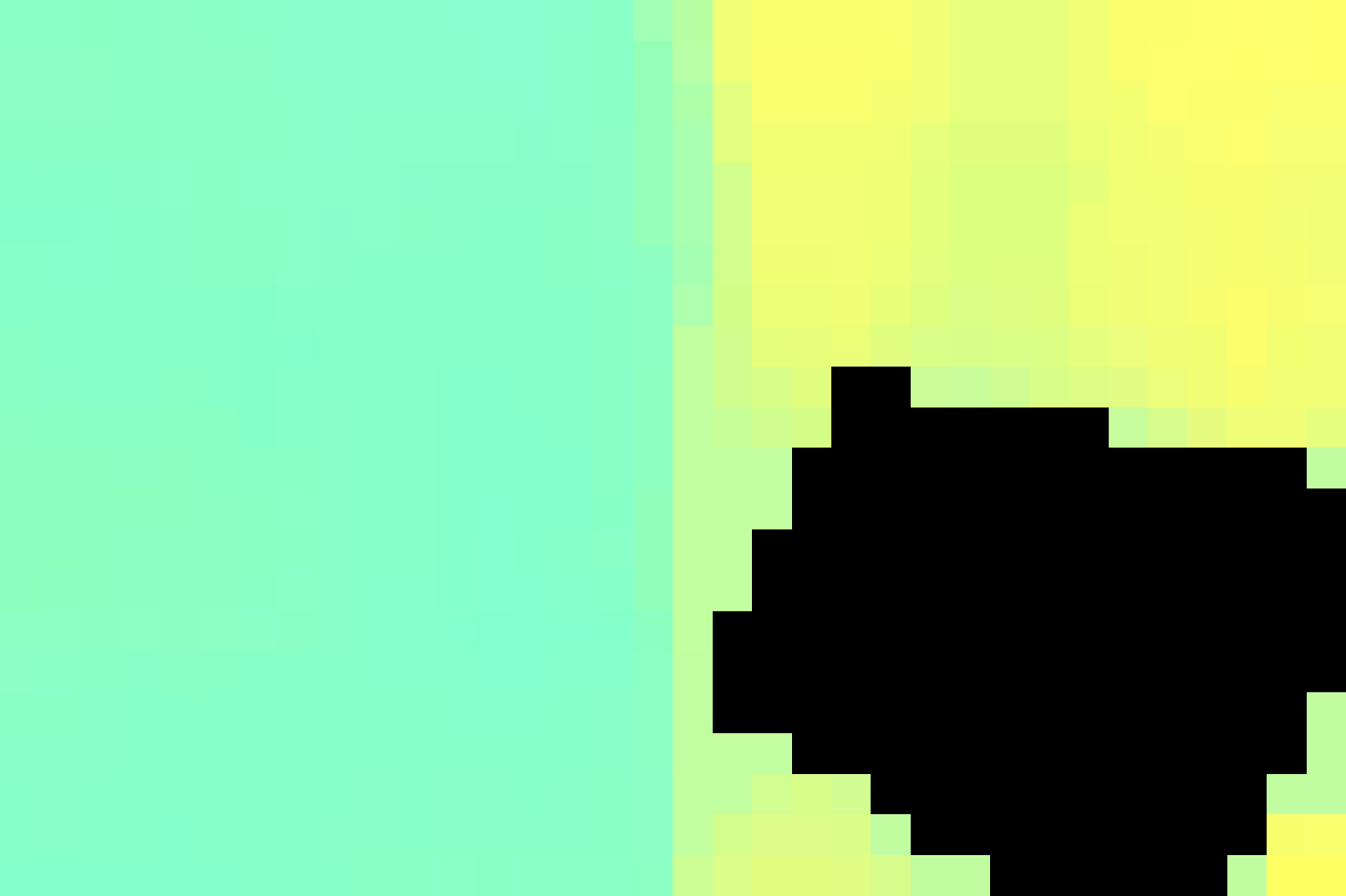}  & \includegraphics[width=0.46\columnwidth]{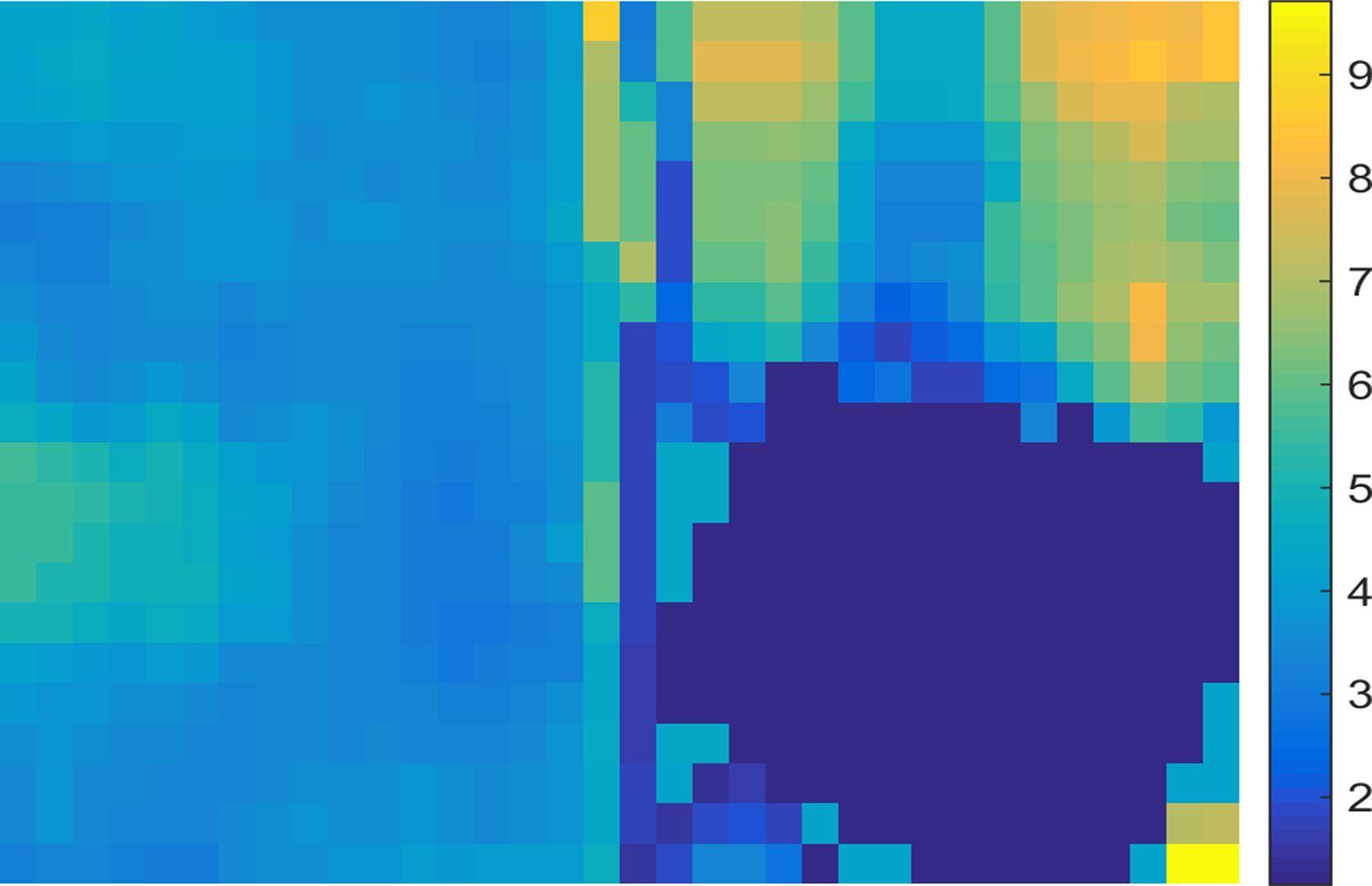} \\
\includegraphics[width=0.45\columnwidth]{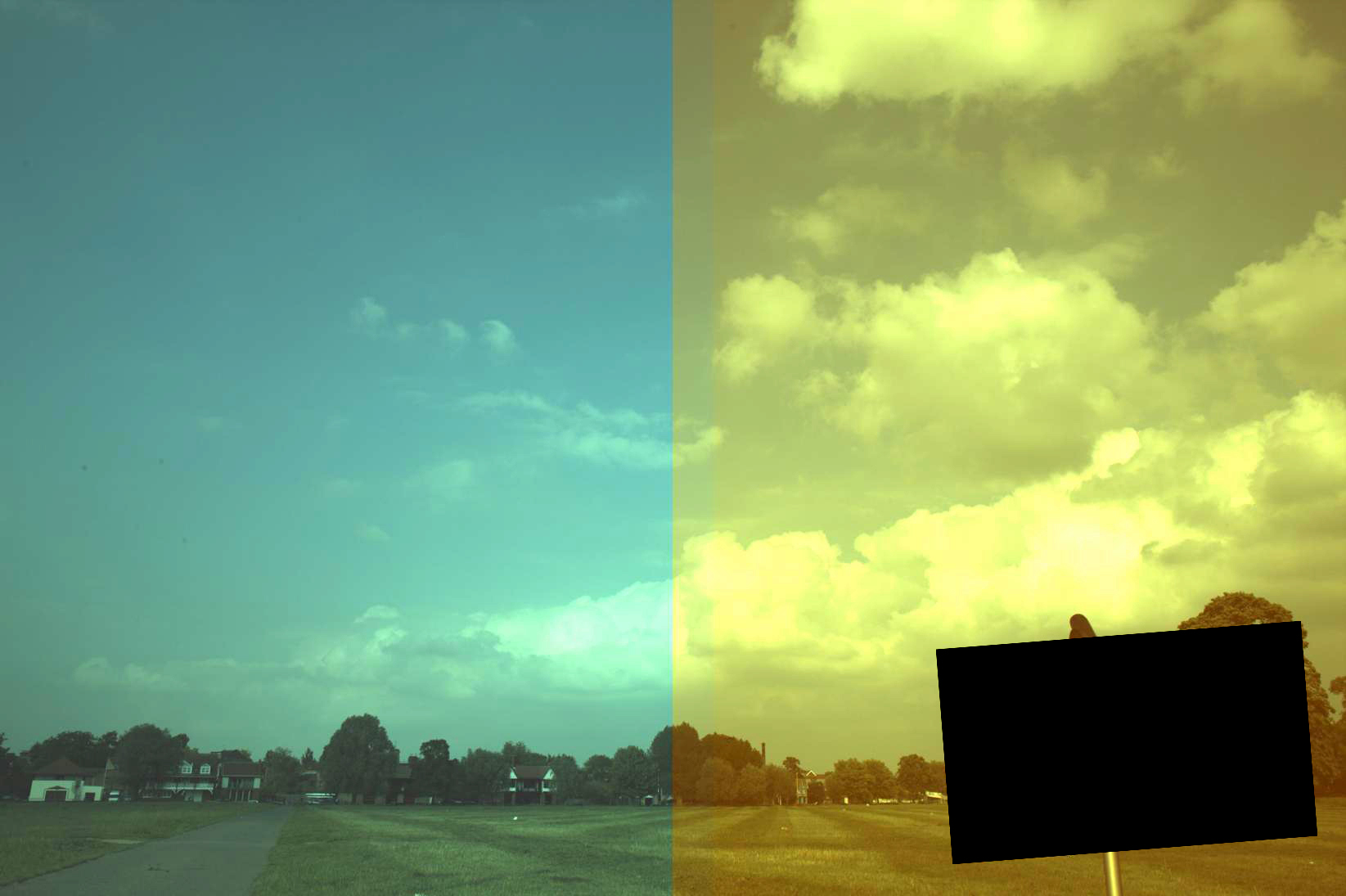}			 & \includegraphics[width=0.45\columnwidth]{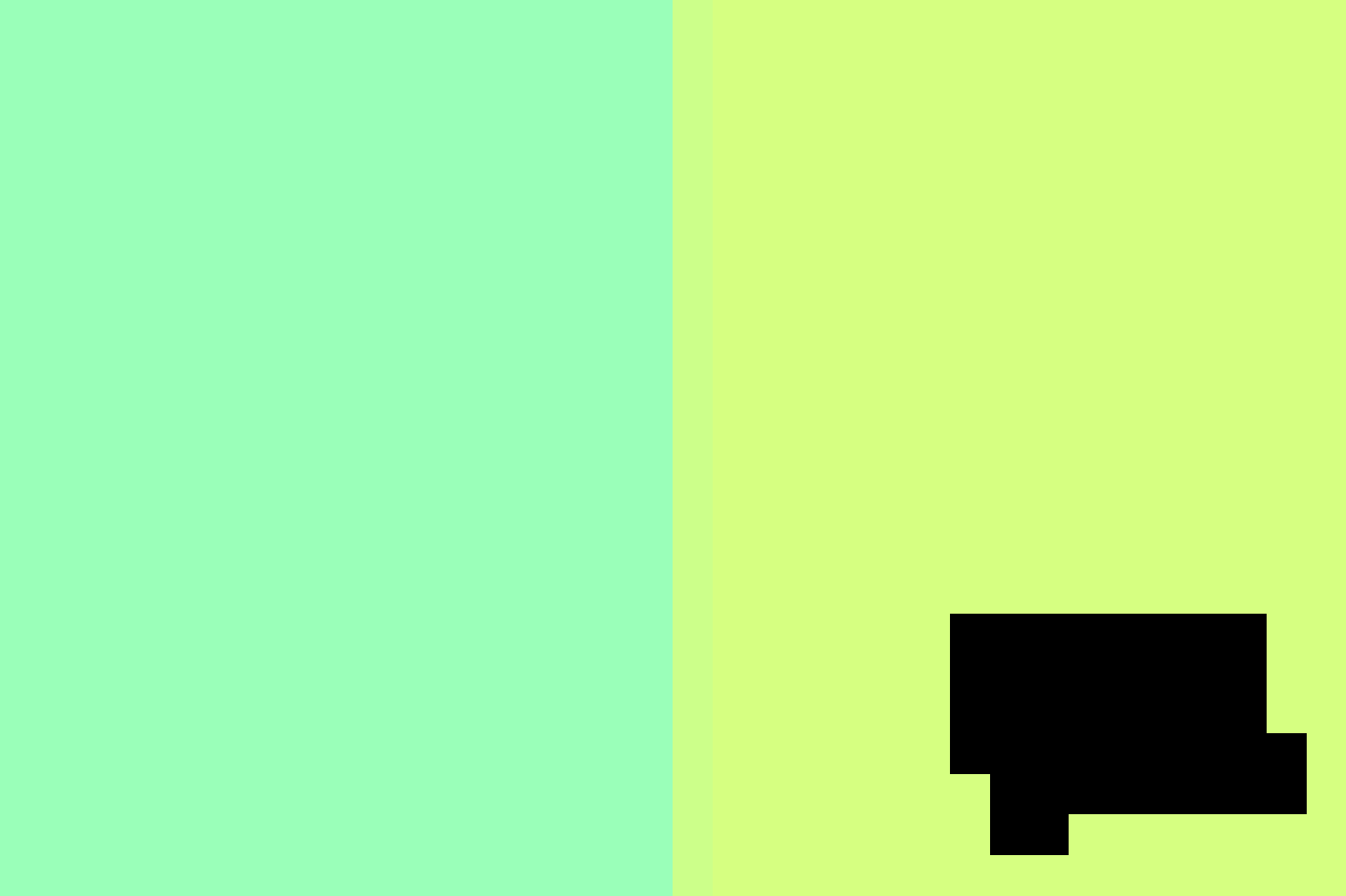} & 
\includegraphics[width=0.45\columnwidth]{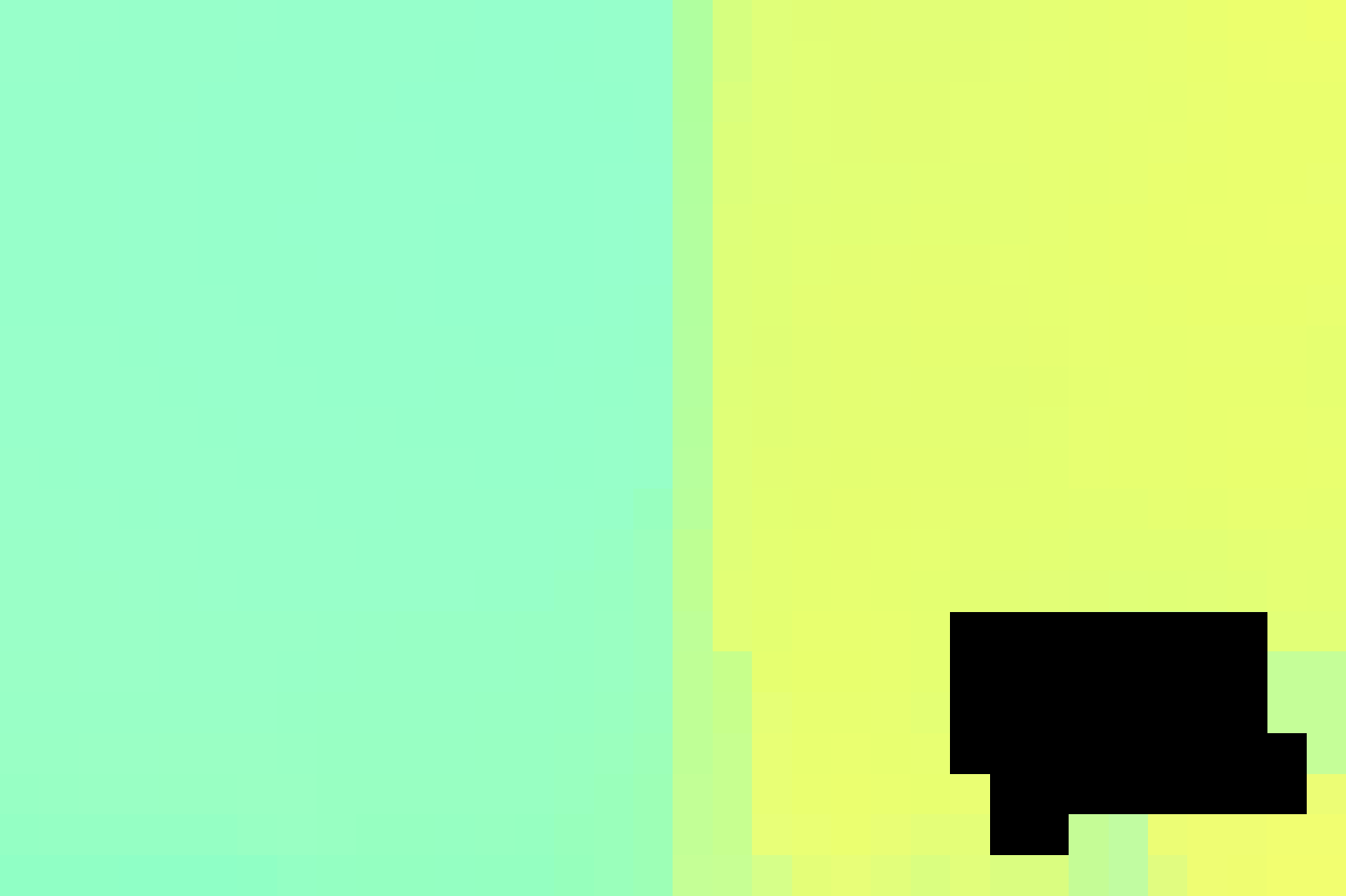} & \includegraphics[width=0.46\columnwidth]{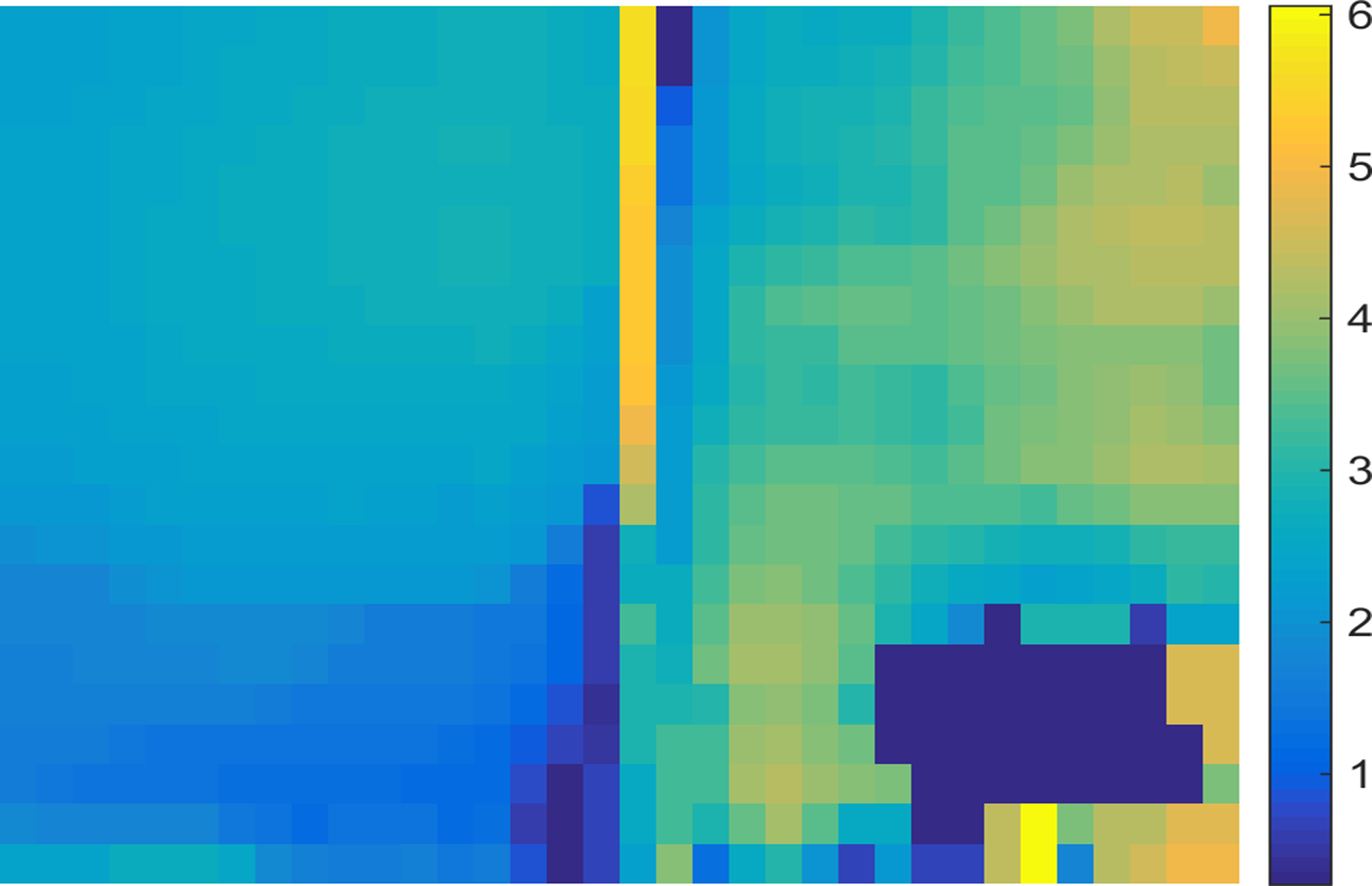}\\
\end{tabular}
	\caption{Examples of local illumination estimation. Left to right: original image, groundruth illumination, local estimation, local angular error map.}
	\label{fig:examplesSV}
\end{figure*}

The numerical results are reported in Table \ref{tab:erroriSV}, while a couple of examples are given in Figure \ref{fig:examplesSV}. For the proposed approach we report three different entries: the first one is the error on each patch; the second and third ones are the patch-by-patch errors by taking into account the spatial arrangement of the patches to perform a spatial filtering of the local estimates: the former employs a 3x3 Gaussian filter, the latter a 3x3 median filter.

\begin{table*}[!ht]
\caption{Angular error statistics obtained on the synthetic RAW dataset of images with spatially varying illumination.}
\label{tab:erroriSV}
\centering
\resizebox{1.25\columnwidth}{!} {
\begin{tabular}{lrrrrrr}
  \toprule
	Algorithm & Min & 10$^{th}$prc &  Med &  Avg & 90$^{th}$prc &  Max  \\
  \midrule
DN   												& 5.90 &  9.99 &  13.38 &  13.62 &  17.09 &  27.71  \\
MLS+GW    									& 0.12 &   3.35 &   8.03 &   8.72 &  14.66 &  32.98  \\
MLS+WP   										& 0.23 &   2.59 &   6.09 &   7.03 &  13.15 &  33.58  \\
\midrule
CNN per patch				 				& 0.00  & 1.01  & 2.83  & 3.72 &  7.78 & 31.78 \\
CNN	gaussian filtering				 & 0.01  & 0.99  & 2.71  & 3.50 &  7.16 & 27.03 \\ 
CNN median filtering					 & 0.01  &  0.96 &    2.66 &    3.39 &    6.88 &   23.13 \\
\bottomrule
\end{tabular}
}
\end{table*} 

\section{Conclusions} 
In this work we have developed a CNN for color constancy. Our algorithm combines feature learning
and regression as a complete optimization process,
which enables us to employ modern training techniques to
boost performance. The experimental results showed that our algorithm  achieves state of the art performance on a standard dataset of RAW images outperforming 21 algorithms in the state-of-the-art belonging to both statistic-based and learning-based classes. Furthermore, a preliminary test, shows that our algorithm can be adapted to estimate local illuminants.

As future work we plan to investigate other pooling strategies to combine patch-based illuminant estimations into a global one. We plan also to investigate if additional information can be fed to the CNN to further improve the performance. We will also conduct a more thorough study about the extension of the proposed approach to local illuminant estimation, conducting the experiments on larger datasets and comparing with more algorithms in the state-of-the-art.

{\small
\bibliographystyle{ieee}
\bibliography{colorConstancyCNN_CVPRW_arxiv}
}

\end{document}